\documentclass[conference]{IEEEtran}
\IEEEoverridecommandlockouts

 \usepackage{amsmath,amssymb,amsfonts}
\usepackage{amsthm}
\usepackage{graphicx}
\usepackage{textcomp}
\usepackage{xcolor}
\usepackage{float}
\usepackage{url}
\usepackage[algo2e,ruled]{algorithm2e}%

\usepackage{paralist} 
\usepackage{listings} 
\usepackage{bbding} 
\usepackage[subrefformat=parens,labelformat=parens]{subfig}
    
\usepackage{pifont}
\usepackage{balance}
\usepackage{booktabs}
\usepackage{enumitem}
\usepackage{diagbox}
\usepackage{bbm}
\newtheorem{definition}{Definition}
\newtheorem{lemma}{Lemma}

\newtheorem{insight}{Insight}
\newtheorem{corollary}{Corollary}
\newcommand{\hide}[1]{}

\newcommand{\bit}{\begin{compactitem}}
\newcommand{\eit}{\end{compactitem}}
\newcommand{\ben}{\begin{compactenum}}
\newcommand{\een}{\end{compactenum}}
\newcommand{\myunderline}[1]{{\underline{#1}}}

\newcommand{\method}{{\sc gen\textsuperscript{2}Out}\xspace}
\newcommand{\methodp}{{\sc gen\textsuperscript{2}Out\textsubscript{0}}\xspace}

\newcommand{\scale}{{Scalable}\xspace}
\newcommand{\effective}{{Effective}\xspace}

\newcommand{\Universal}   {{Principled and Sound}\xspace}

\newcounter{x}
\newtheorem{inneraxiom}{{Axiom}}
\newenvironment{axiom}[1]
  {\renewcommand\theinneraxiom{A\arabic{x} (#1)}\inneraxiom\stepcounter{x}}
  {\endinneraxiom}

\newcommand{\ifr}{{\sc iF}\xspace}

\newcommand{\rcf}{{\sc RRCF}\xspace}
\newcommand{\loda}{{\sc LODA}\xspace}
\newcommand{\pca}{{\sc PCA}\xspace}
\newcommand{\lof}{{\sc LOF}\xspace}

\newcommand{\red}[1]{\colorbox{red!50}{#1}}

\newcommand{\blue}[1]{\colorbox{blue!40}{#1}}
\newcommand{\green}[1]{\colorbox{green!40}{#1}}

\newcommand{\bs}[1]{\boldsymbol{#1}}

\newcommand{\tree}{\textsc{atomicTree}\xspace}

\newcommand{\cmark}{\ding{51}}

\newcommand{\numdatasets}{{26\xspace}}

\newcommand{\linearlemma}{{Power Depth Property}\xspace}
\newcommand{\llshort}{{PDP}\xspace}

\newcommand{\dlimit}{$d_{limit}$} 
\newcommand{\leaf}{$l_{busy}$} 

\newtheorem{iprob}{Informal Problem}

\SetCommentSty{mycommfont}

\newcommand{\fctree}{\textsc{fixedCutTree}\xspace}
\newcommand{\group}   {{group}\xspace}
\newcommand{\dgen}   {{doubly general}\xspace}
\newcommand{\genanomaly}   {{generalized anomaly}\xspace}
\newcommand{\genanomalies}   {{generalized anomalies}\xspace}
\newcommand{\algo}   {{Doubly-general}\xspace}
\newcommand{\xray}   {{\sc{X-ray}}\xspace}
\newcommand{\sxray}   {{\sc{Apex}}\xspace}
\newcommand{\point}{point\xspace}
\newcommand{\ptanomaly}   {{point-anomaly}\xspace}
\newcommand{\ptanomalies}   {{point-anomalies}\xspace}
\newcommand{\grpanomalies}   {{group-anomalies}\xspace}

\newcommand{\http}{{\texttt{http}}\xspace}
\newcommand{\eeg}{{\texttt{EEG}}\xspace}
\newcommand{\optdigits}{{\texttt{optdigits}}\xspace}

\newcommand{\rulesep}{\unskip\ \vrule\ }    

\begin{document}

\title{\method: Detecting and Ranking Generalized Anomalies
}

\makeatletter
\newcommand{\newlineauthors}{%
  \end{@IEEEauthorhalign}\hfill\mbox{}\par
  \mbox{}\hfill\begin{@IEEEauthorhalign}
}
\makeatother

\author{
\IEEEauthorblockN{Meng-Chieh Lee*}
\IEEEauthorblockA{Carnegie Mellon University \\
mengchil@cs.cmu.edu}
\and
\IEEEauthorblockN{Shubhranshu Shekhar*}
\IEEEauthorblockA{Carnegie Mellon University \\
shubhras@andrew.cmu.edu}
\newlineauthors
\IEEEauthorblockN{Christos Faloutsos}
\IEEEauthorblockA{Carnegie Mellon University \\
christos@cs.cmu.edu}
\and
\IEEEauthorblockN{T. Noah Hutson}
\IEEEauthorblockA{Barrow Neurological Institute \\
timothy.hutson@commonspirit.org}
\and
\IEEEauthorblockN{Leon Iasemidis}
\IEEEauthorblockA{Barrow Neurological Institute \\
leonidas.jassemidis@commonspirit.org}
\thanks{* Both authors contributed equally to this work.}
}

\IEEEpubid{\makebox[\columnwidth]{978-1-6654-3902-2/21/\$31.00~\copyright2021 IEEE \hfill} \hspace{\columnsep}\makebox[\columnwidth]{ }}

\maketitle

\begin{abstract}
In a cloud of $m$-dimensional data points, how would we spot,
as well as rank, both
 single-point- as well as \group- anomalies? 
 We are the first to 
 generalize anomaly detection in two dimensions:
  The first dimension is that we handle
  both point-anomalies, as well as \group-anomalies,
  under a  unified view - we shall refer to them as
  {\em \genanomalies}. 
 The second dimension is that \method not only detects,
 but also ranks, anomalies in suspiciousness order.
Detection, and ranking, 
of  anomalies 
has numerous applications:
For example, in EEG recordings of an epileptic patient, 
an anomaly may indicate a seizure;
in computer network traffic data, it may signify a power failure,
or a DoS/DDoS attack.

We start by setting some reasonable axioms;
surprisingly, none of the earlier methods pass all the axioms.
Our main contribution is
the \method algorithm, that 
has the following desirable properties: (a) 
{\em \Universal} anomaly scoring that obeys the axioms for detectors, 
(b) {\em \algo} in that it detects, as well as ranks \genanomaly -- both \point- and \grpanomalies,
(c) {\em \scale}, it is fast and scalable,
linear on input size.
(d) {\em \effective}, experiments on real-world epileptic recordings (\myunderline{\textit{200GB}}) 
demonstrate effectiveness of \method as confirmed by clinicians. 
Experiments on \myunderline{\textit{27}} real-world benchmark datasets show that \method detects ground truth \group{s},
matches or outperforms \ptanomaly baseline algorithms
on accuracy, with no competition for group-anomalies
and requires
about \myunderline{\textit{2 minutes for 1 million data \point{s}}}
on a stock machine.
\end{abstract}

\section{Introduction}
\label{sec:intro}
How would we spot and rank
single-point- as well as \grpanomalies? How can we draw attention of the clinician to strange brain activities in multivariate EEG recordings of an epileptic patient?
How could we design an anomaly score function, 
so that it assigns intuitive scores to both \point{-}, as well as \grpanomalies?
Our goal is to design a principled and fast anomaly detection algorithm for a given cloud of $m$-dimensional point-cloud data that provides a unified view as well as a scoring function for each \genanomaly. This has numerous applications (intrusion detection in computer networks, automobile traffic analysis, outlier\footnote{We use outlier and anomaly interchangeably in this work.} detection in a collection of feature vectors from, say, medical images, or twitter users, or DNA strings, and more).

Our motivating application is seizure detection in EEG recordings.
Specifically, we want to spot those parts of the brain,
and those time-ticks, that a seizure happened.
 Epilepsy is a neurodegenerative disease that affects $1-2$\% of the world’s population and is characterized by recurrent seizures that intermittently disrupt the normal function of the brain through paroxysmal electrical discharges~\cite{shorvon2009epilepsy}. At least $30$\% of patients with medically refractory epilepsy are resistant to the mainstay treatment by anti-epileptic drugs (AEDs). These patients may benefit from surgical therapy. A significant challenge of this therapy is identification of the region of the brain where seizures are originating, that is, the epileptogenic focus~\cite{krishnan2015epileptic, vlachos2016concept}.  This region is then surgically either resected or electrically stimulated over time to control upcoming seizures long prior to their occurrence~\cite{tsakalis2006control, chakravarthy2009controlling, hutson2018predictability}. Accurate identification of the epileptogenic focus is therefore of high significance for the treatment of epilepsy.
\begin{figure*}[]
{
\centering
\subfloat[\label{fig:crown2} \eeg data]
{\includegraphics[scale=0.75]{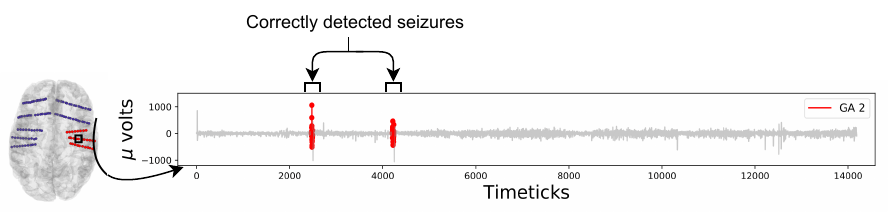}}
\hspace{3mm}
\rulesep
\hspace{3mm}
\subfloat[\label{fig:crown:http}Heatmap of \http data]
{\includegraphics[scale=0.09]{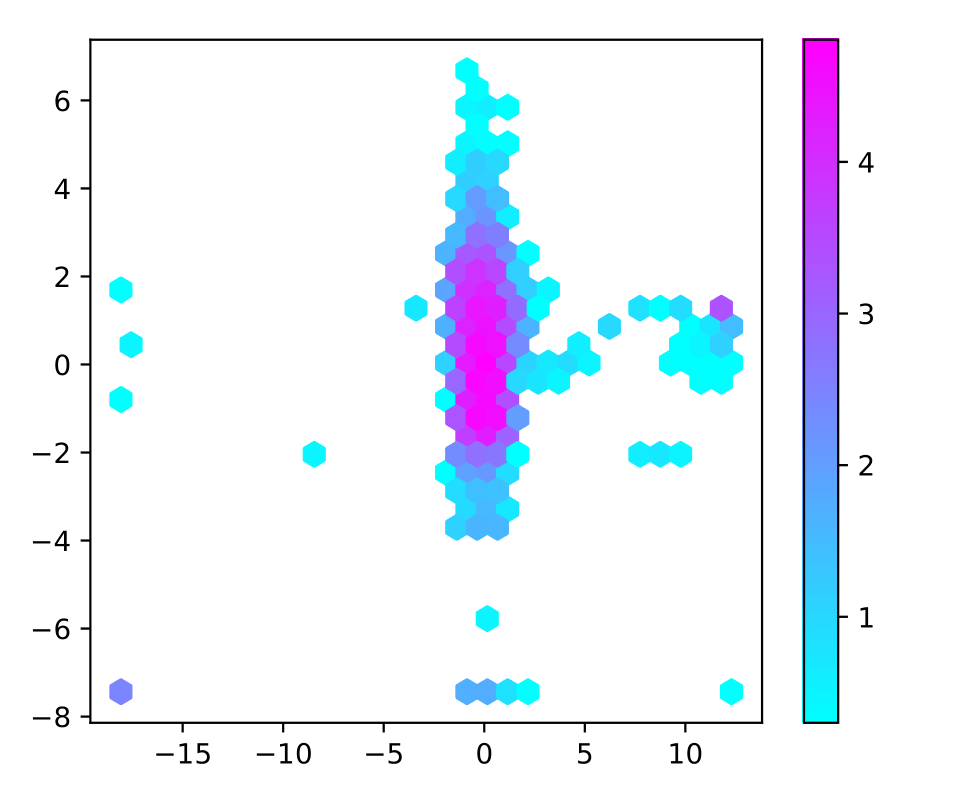}}
\subfloat[\label{fig:crown1} Detected \grpanomalies]
{\includegraphics[scale=0.042]{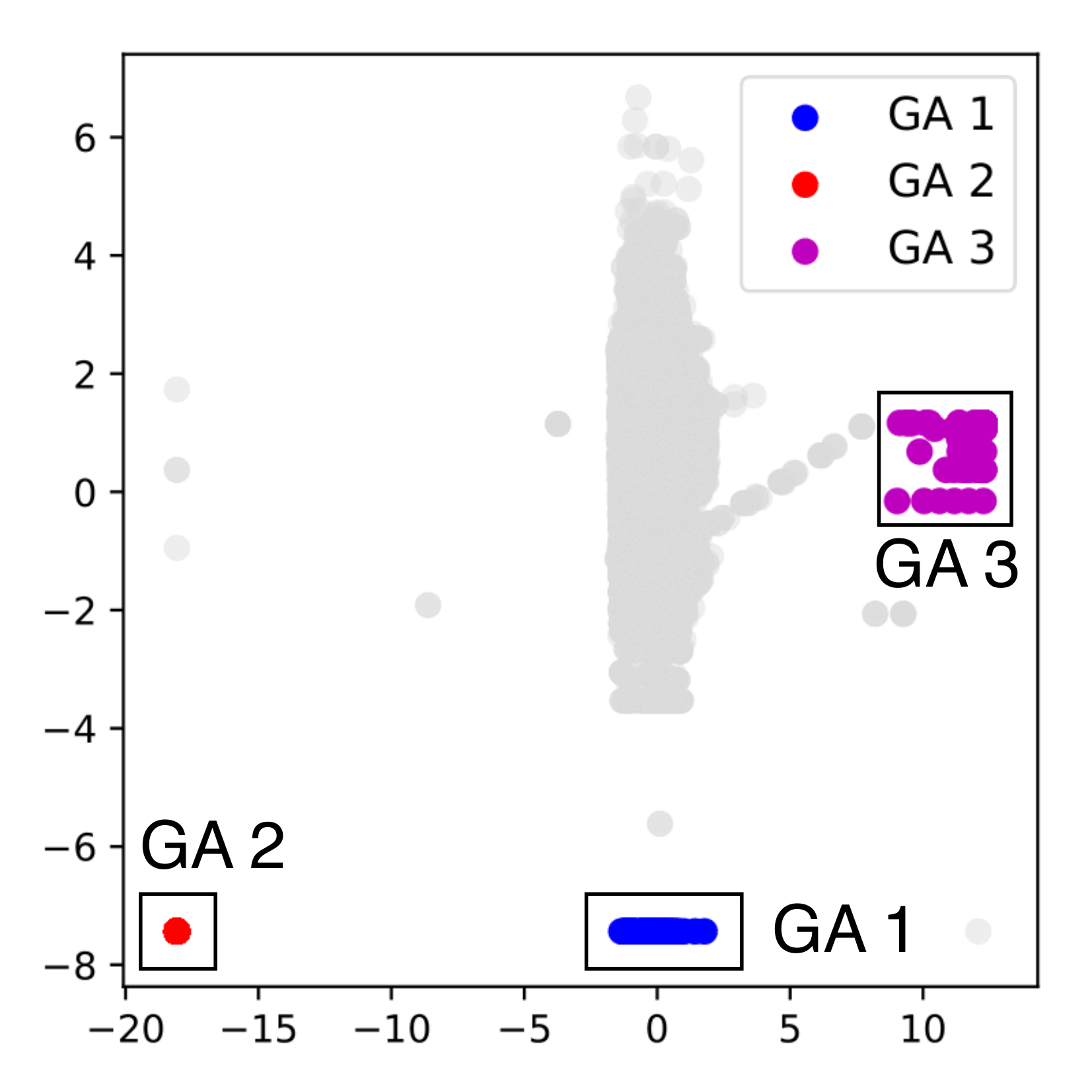}}
}

\caption{\label{fig:crown}
	(a)  \myunderline{\method matches ground truth.} Brain scan of the patient with electrode positions (\emph{left}), and detected \group{s} shown in color red~(\emph{right}), that matches the ground truth seizure locations.
	(b) Heatmap of {\em http} intrusion detection dataset (c) \myunderline{\method correctly spots \group (DDoS) attacks} in the intrusion detection dataset, marked GA1, GA2 and GA3.
	\looseness=-1}
\end{figure*}

As suggested by the application domain, to achieve better outcomes for patients, it is critical to direct attention of the clinician to the anomalous time periods in the brain activity in order of their suspiciousness. 
The problem is two-fold: (a) {\em detection}, as well as (b) {\em ranking} of \genanomalies.
We want a scoring function for \genanomalies,
such that in
the EEG/epilepsy setting it would score the \group{s} which may correspond to anomalous periods e.g. seizure
and draw attention to most anomalous time periods; thus aiding a domain expert in decision making. As we show in Section~\ref{subsec:axioms}, 
we propose some intuitive axioms, that
a \genanomaly detector should obey.


\begin{iprob}
[Doubly-general anomaly problem]\hfill
\begin{itemize}
\item {\bfseries Given} a point-cloud dataset from an application setting, 
\item {\bfseries find} the \ptanomalies and \grpanomalies, and
\item {\bfseries rank} them in suspiciousness order.

	\end{itemize}
\end{iprob}


\noindent {\bf Generality of approach:}
In most machine learning (ML) algorithms, we operate on clouds of points (after embedding, 
after auto-encoding etc).
For example, time series is transformed into some form of $m-$dimensional cloud~\cite{blazquez2021review} for further analysis;
in images, numerical features are generated for learning tasks e.g. Imagenet~\cite{krizhevsky2012imagenet}. 
Thus, the proposed approach can be applied to diverse real data including point cloud, time-series and image data.\looseness=-1

Figure~\ref{fig:crown} illustrates the effectiveness of our method -- \method detects \grpanomalies that correspond to seizure period in the patient; and, detects DoS/DDoS attack as \grpanomalies.

We propose \method, which 
has
the following properties:

\bit
\item {\bf \Universal:} We identify five axioms
(see Section \S\ref{subsec:axioms}) 
and show that the proposed
 \method obeys them, in contrast to top competitors.

\item {\bf \algo:} \method is \dgen. First dimension of generalization is size of anomalies -- detecting \point- and \grpanomalies. Second dimension of generalization is scoring and ranking of \genanomalies -- both \point- and \grpanomalies. 


	
\item 
{\bf \scale:}
Linear on the input size (see Figure~\ref{fig:gen:scale}).
\item{\bf \effective}:
Applied on real-world data~(see Figure~\ref{fig:crown} and \ref{fig:aproc}),
\method wins in most cases over benchmark datasets for \point anomaly detection. For \group anomaly detection, \method has no competitors as they need group structure information, and it agrees with ground truth on seizure detection.

\eit

\noindent{\bf Reproducibility:} Our source code and public datasets are at \url{https://github.com/mengchillee/gen2Out}. Our epilepsy EEG dataset involves real patients and  requires NDA.

\begin{table*}[]
		\centering
	\caption{\underline{\method matches all the specs}. Qualitative comparison of \method against top competitors showing that every competitor misses one or more features. \label{tab:salesman}}
	\resizebox{0.75\linewidth}{!}{
		\begin{tabular}{l| c c c c c c c | c}
			\diagbox{Property}{Method}& \rotatebox{90}{LODA~\cite{pevny2016loda}} & \rotatebox{90}{RRCF~\cite{guha2016robust}} & \rotatebox{90}{IF~\cite{liu2008isolation}} & \rotatebox{90}{OCSMM\cite{muandet2013one}}& \rotatebox{90}{AAE-VAE\cite{chalapathy2018group}} & \rotatebox{90}{MGM\cite{xiong2011hierarchical}} & \rotatebox{90}{GLAD \cite{yu2015glad}}& \rotatebox{90}{\method} \\
			\midrule
			Obeys Axioms (see \S\ref{subsec:axioms})&   &  &  & \textbf{?} &\textbf{?}  &\textbf{?} &\textbf{?} &\blue{\cmark}\\
			Discover point anomalies&  \cmark & \cmark & \cmark & &&& &\blue{\cmark}\\
			Rank point anomalies& \cmark  & \cmark & \cmark & &&& &\blue{\cmark}\\
			Discover \group anomalies&   &  &  & &&&  \textbf{?} &\blue{\cmark}\\
			Rank \group anomalies&   &  &  &\cmark &\cmark&\cmark& \cmark &\blue{\cmark}\\
			Jointly rank point- and \group{-} anomalies&   &  &  & &&& &\blue{\cmark}\\
			Scalable & \cmark & \textbf{?} & \cmark & \textbf{?}& \cmark& \textbf{?}& \textbf{?}&\blue{\cmark}\\
		\end{tabular}
	}  
\end{table*}

\section{Background and Related Work}
\label{sec:background}

Anomaly detection is a well-studied problem. Recent works~\cite{boukerche2020outlier, chandola2009anomaly, aggarwal2015outlier, gupta2013outlier, toth2018group} provide a detailed review of many methods for anomaly and outlier detection. 
As shown in Table~\ref{tab:salesman}, \method is the only method that matches the specs.
Here, we review anomaly detection methods for point- and -\group{-} anomalies.

{\em Point Anomaly Detection.} 
Model-based and density-based methods for outlier detection are quite popular for point cloud data. Principal component analysis (\pca) based detectors~\cite{shyu2003novel} assume that the data follows a multi-variate normal distribution. Local outlier factor (\lof)~\cite{breunig2000lof} flags instances that lie in low-density regions. 
Clustering based methods~\cite{he2003discovering} score instances or small clusters by their distance to large clusters. However, these methods suffer from too many false positives as they are not optimized for detection~\cite{liu2012isolation}.
Recently, a surge of focus has been on ensemble-based detectors that have been shown to outperform base detectors and are considered state-of-the-art for outlier detection~\cite{emmott2013systematic}. Isolation forest~\cite{liu2008isolation} (\ifr), a state-of-the-art ensemble technique, builds a set of randomized trees that allows approximating the density of instances in a random feature subspace. Emmott et al. ~\cite{emmott2013systematic}(2015) show that \ifr significantly outperforms other detectors such as \lof. \ifr~\cite{liu2008isolation} shows that \lof has a high computation complexity (quadratic) and does not scale for large datasets. After that two more methods \loda~\cite{pevny2016loda} and Random Cut Forests (\rcf)~\cite{guha2016robust} have been proposed as state-of-the-art methods. \loda is projection-based histogram ensemble that works well in many real settings. \rcf improve upon \ifr and use a data sketch that preserves pairwise distances.


{\em Group Anomaly Detection.} Numerous methods have been proposed for \group anomaly detection~\cite{muandet2013one, chalapathy2018group, xiong2011hierarchical, yu2015glad}. 
Earlier approaches~\cite{muandet2013one, chalapathy2018group, xiong2011hierarchical} require the group memberships of the points known apriori, while 
Yu et al.~\cite{yu2015glad} requires information on pairwise relations among data points.
Moreover, these methods focus only on scoring \grpanomalies, and ignore \ptanomalies unlike our method.
\method detects and ranks anomalous \point{s} and \group{s}, without requiring additional information on group structure of the dataset.
As mentioned above, Table~\ref{tab:salesman} summarizes comparison of \method against state-of-the-art \point- and \group- anomaly detection methods. As such none of the methods has all the features of Table~\ref{tab:salesman} .

{\em Fractals and multifractals}: In order to stress test our method, we use self-similar (fractals) clouds of points.
We created the fractal images (Sierpinski triangle, biased line and `fern' etc.), using the method and the code from Barnsley and Sloan~\cite{10.5555/44935.44950}. We used the `uniform' version (that is, for  the Sierpinski triangle, all the miniature versions have the same weight of 1/3), also generated the `biased' version of triangle using weights (0.6, 0.3, 0.1), and `biased line' with \emph{bias} $b=0.8$ using weights (0.8, 0.2) that is $b$ of the data points go to the first half of the line, and in this half, $b$ of the data points go to first quarter of the line, and so on recursively (this, informally, is the 80-20 law).

\section{Proposed Axioms and Insights}
\label{sec:preliminaries}
\begin{figure*}[t]
	\centering
	\subfloat[A1: Distance Axiom]
	{\label{fig:a1}\includegraphics[scale=0.3]{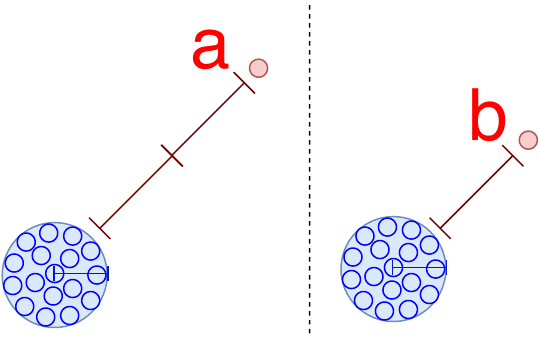}}
	\hspace{5mm}
	\subfloat[A2: Density Axiom]
	{\label{fig:a2}\includegraphics[scale=0.33]{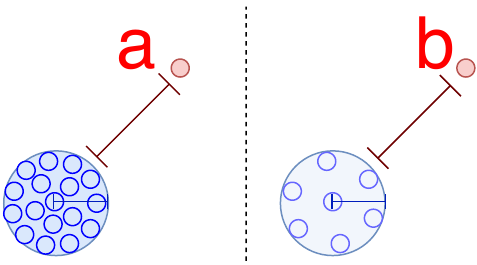}}
	\hspace{5mm}
	\subfloat[A3: Radius Axiom]
	{\label{fig:a3}\includegraphics[scale=0.28]{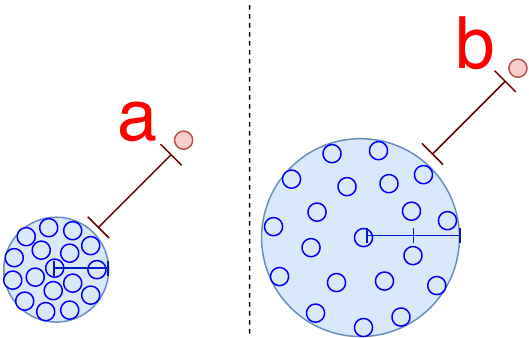}}
	\hspace{5mm}
	\subfloat[A4: Angle Axiom]
	{\label{fig:a4}\includegraphics[scale=0.22]{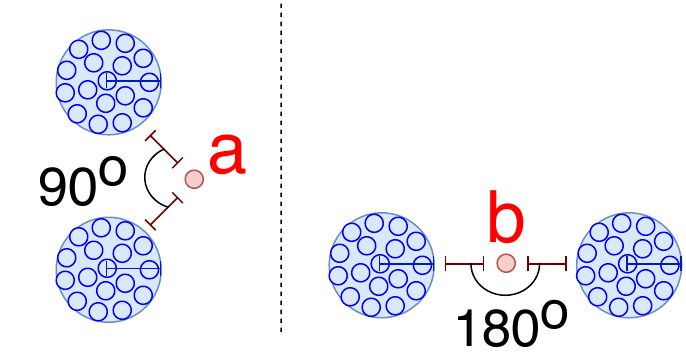}}
	\hspace{5mm}
	\subfloat[A5: Group Axiom]
	{\label{fig:a5}\includegraphics[scale=0.26]{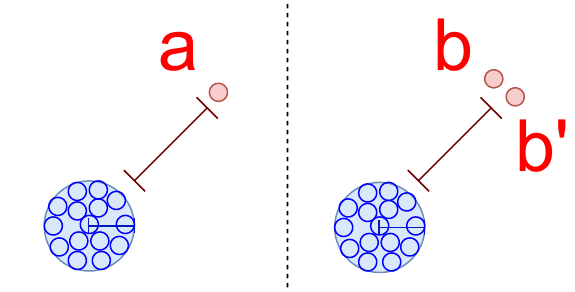}}
	\caption{Illustration of Axioms\label{fig:axiom}}
	\vspace*{-0.1in}
\end{figure*}

In this section, we explain our proposed axioms in detail and give the insights. It is worth noting that these axioms are proposed to examine whether an anomaly detector is provided with the ability to compare the scores across datasets. The assumption is that, there are two different datasets with the same application setting. Although some of the axioms seem to be popular in single dataset setting, they are not considered and even ever mentioned by other studies when there is more than one dataset. The observed insights are critical and penetrate this research. These greatly inspire us on selecting the core part of our anomaly detector.

\subsection{Proposed Axioms}
\label{subsec:axioms}
We propose five  axioms an ideal anomaly detector should follow: producing higher anomaly scores when an instance is farther away from data kernel ({\emph distance aware}), or lies in low density locality ({\em density, radius and group aware}), and not aligned with majority of data ({\emph angle aware}~\cite{kriegel2008angle}). In the following, let $\boldsymbol{a} \in \mathrm{R}^m$ and $\boldsymbol{b} \in \mathrm{R}^m$ be $m-$dimensional anomalies in point cloud datasets $S_a$ and $S_b$ respectively. Additionally, suppose that normal observations are distributed uniformly in a disc in the datasets as shown in Figure~\ref{fig:axiom} and $s(.)$ is the \genanomaly score function.
{\parindent0pt 
\begin{axiom}{Distance Aware}
    {\em  All else being equal, the farther point from the normal observations is more anomalous.}
    \begin{equation*}
    \left .
    \begin{array}{l}
    S_a - \{\bs{a}\}= S_b - \{\bs{b}\},\\
    dist(\bs{a}, S_a) > dist(\bs{b}, S_b)\\
    \end{array} \right\rbrace   
    \implies  s(\bs{a}) > s(\bs{b})
    \end{equation*}
\end{axiom}
\vspace{-0.1in}
\begin{axiom}{Density Aware}
    {\em All else being equal, denser the cluster of points, more anomalous the outlier.}
    \begin{equation*}
    	\left .
    	\begin{array}{l}
	    	dist(\bs{a}, S_a) = dist(\bs{b}, S_b),\\
	    	density(S_a) >density(S_b)\\    	
    	\end{array} \right\rbrace   
        \implies  s(\bs{a}) > s(\bs{b})
    \end{equation*}
\end{axiom}
\begin{axiom}{Radius Aware}
    {\em All else being equal, for a given number of observations, smaller the radius of the cluster of points, more anomalous the outlier.}
    \begin{equation*}
    	\left .
    	\begin{array}{l}
        |S_a| = |S_b|, \\
        dist(\bs{a}, S_a) = dist(\bs{b}, S_b),\\
        radius(S_a) < radius(S_b)
        \end{array} \right\rbrace   
        \implies s(\bs{a}) > s(\bs{b})
    \end{equation*}
\end{axiom}
\begin{axiom}{Angle Aware}
    {\em All else being equal, smaller the angle of a point with respect to cluster of observation, more anomalous the outlier.}
    \begin{equation*}
    \left .
    	\begin{array}{l}
	        |S_a| = |S_b|, \\
	        density(S_a) = density(S_b),\\
	        radius(S_a) = radius(S_b),\\
	        angle(\bs{a}, S_a) < angle(\bs{b}, S_b)\\
        \end{array}\right\rbrace   
        \implies  s(\bs{a}) > s(\bs{b})
    \end{equation*}
\end{axiom}

\begin{axiom}{Group-size Aware}
	{\em All else being equal,  the least populous \group , the more anomalous it is.}
	$$\text{Let } g_a \subset S_a, g_b \subset S_b \text{ are the \group{s}}.$$
	\begin{equation*}
		\left .
		\begin{array}{l}
			|g_a| < |g_b|\\
			|S_a - g_a| = |S_b - g_b|, \\
			density(S_a) = density(S_b),\\
			radius(S_a) = radius(S_b),\\
		\end{array}\right\rbrace   
		\implies  s(g_a) > s(g_b)
	\end{equation*}
\end{axiom}
}

\noindent{\em Justification for Axioms}
Axiom A1 is self explanatory as shown in Figure~\ref{fig:a1}. The outlier point (shown in color red) in the left dataset (Figure~\ref{fig:a1}) being farther from the normal observations should be more anomalous.

Consider the case of social networks. A node reachable via $k$ hops from a close friends group should be more anomalous compared to reachable via $k$ hops from a colleagues group. Figure~\ref{fig:a2} illustrates Axiom A2 where the outlier in the left dataset should be more anomalous.\looseness=-1

As shown in Figure~\ref{fig:a3}, for the same number of observations, the larger radius cluster would have a larger distance among points. Therefore, the outlier in the left dataset with smaller radius should be more anomalous.\looseness=-1

The farther points would tend to form a smaller angle with the cluster of observations~(see Figure~\ref{fig:a4}) and should be more anomalous in the left dataset.

The \group $g_a=\{\bs{a}\}$ consisting of one point Figure~\ref{fig:a5} is intuitively more anomalous compared to \group $g_b = \{\bs{b}, \bs{b}'\}$ containing more data points. For example, if $g_b$ has $1000$ points, it is not an anomaly anymore.

\begin{figure}[]
	\centering
	\includegraphics[scale=0.35]{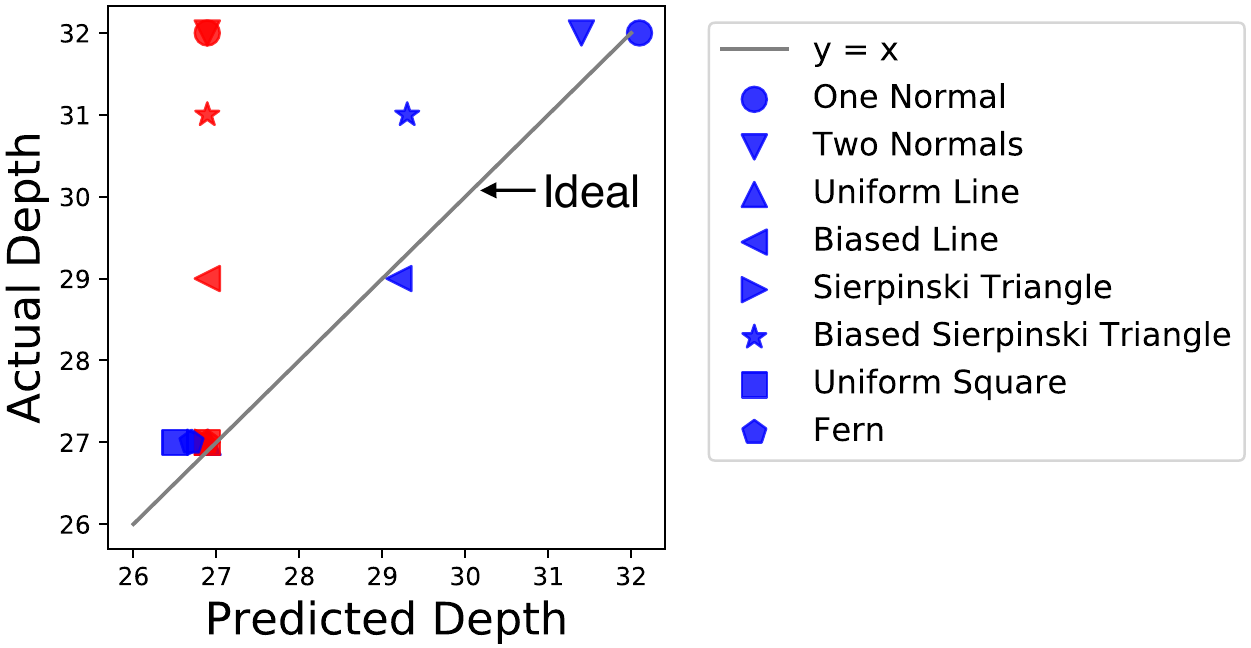}
	\caption{{\myunderline{\method wins}}~(in color blue) as the estimated depth is close to $45^o$ line. \ifr estimates the same depth for each dataset with $\#$samples=1M.}
	\label{fig:dp}
\end{figure}
\subsection{Insights}
\label{subsec:insights}
In this section, we are given the observations $X = \{ \bs{x_1}, \dots, \bs{x_n} \}$ where $\bs{x_i} \in \mathrm{R}^m$~(see Table~\ref{tab:symbols} for symbol definitions) for the anomaly detection. Our goal is to design an anomaly detector that obeys the axioms proposed in \S\ref{subsec:axioms}. 
The intuition for the selection of basic model is that, according to the five axioms in Figure~\ref{fig:axiom}, point `a' in the first dataset should always have higher probability to be separated out comparing the point `b' in the second dataset. \tree has the properties which are very close to our demand.
Here, we consider a randomized tree \tree data structure with the following properties -- (i) Each node in the tree is either {\em leaf} node, or an internal node with two children, (ii) internal nodes store an attribute-value pair and dictate tree traversal. Given $X = \{ \bs{x_1}, \dots, \bs{x_n} \}$, \tree is grown through recursive division of $X$ by randomly selecting an attribute and a split value until all the leaf nodes contain exactly one instance (hence the name \tree) of observations assuming that observations are distinct.
We randomly generate more than one tree to build a forest, to reduce the variance and detect outliers in subspaces.

We make a number of interesting observations while 
empirically investigating the process of tree growth for a variety of data distributions including multi-fractals. In Figure~\ref{fig:depthdist}, we report depth (height) distribution of randomized trees averaged over $100$ trees. We sample a number of points ($|X| \in \{2^{10}, 2^{11}, 2^{12}, 2^{20}\}; m=2$) from each data distribution (shown in Figure~\ref{fig:depthdist} (a), (b), (c), (d), (i), (j), (k), (l)) and plot their corresponding depth (height) distribution (shown in Figure~\ref{fig:depthdist} (e), (f), (g), (h), (m), (n), (o), (p)). Notice that the number of points~($2^\text{x}; \text{x}  \in\{6, 7, \dots\}$) in the tree grows linearly with the average depth for any given dataset. In Figure~\ref{fig:dp}, we plot the predicted depth for each of the distributions against the actual depth of the tree for those distributions shown in Figure~\ref{fig:depthdist} by fitting to this linear trend. 
We present the following lemma based on the observations and draw the following insights.\looseness=-1
\begin{figure}[!ht]
	\centering
	\subfloat[One Normal]
	{\label{fig:norm}\includegraphics[scale=0.17]{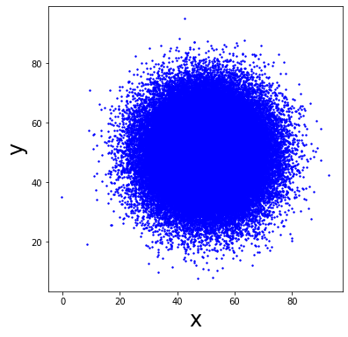}}
	\subfloat[Two Normals]
	{\label{fig:twonorm}\includegraphics[scale=0.17]{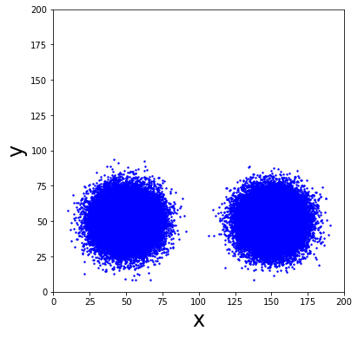}}
	\subfloat[Uniform Line]
	{\label{fig:line}\includegraphics[scale=0.17]{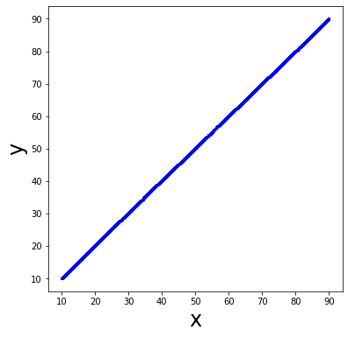}}
	\subfloat[Biased Line]
	{\label{fig:fractalline}\includegraphics[scale=0.17]{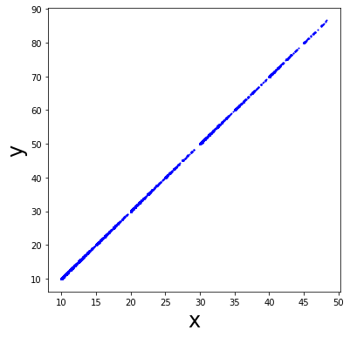}}
	\\
	\subfloat[One Normal]
	{\label{fig:norm_dist}\includegraphics[scale=0.1]{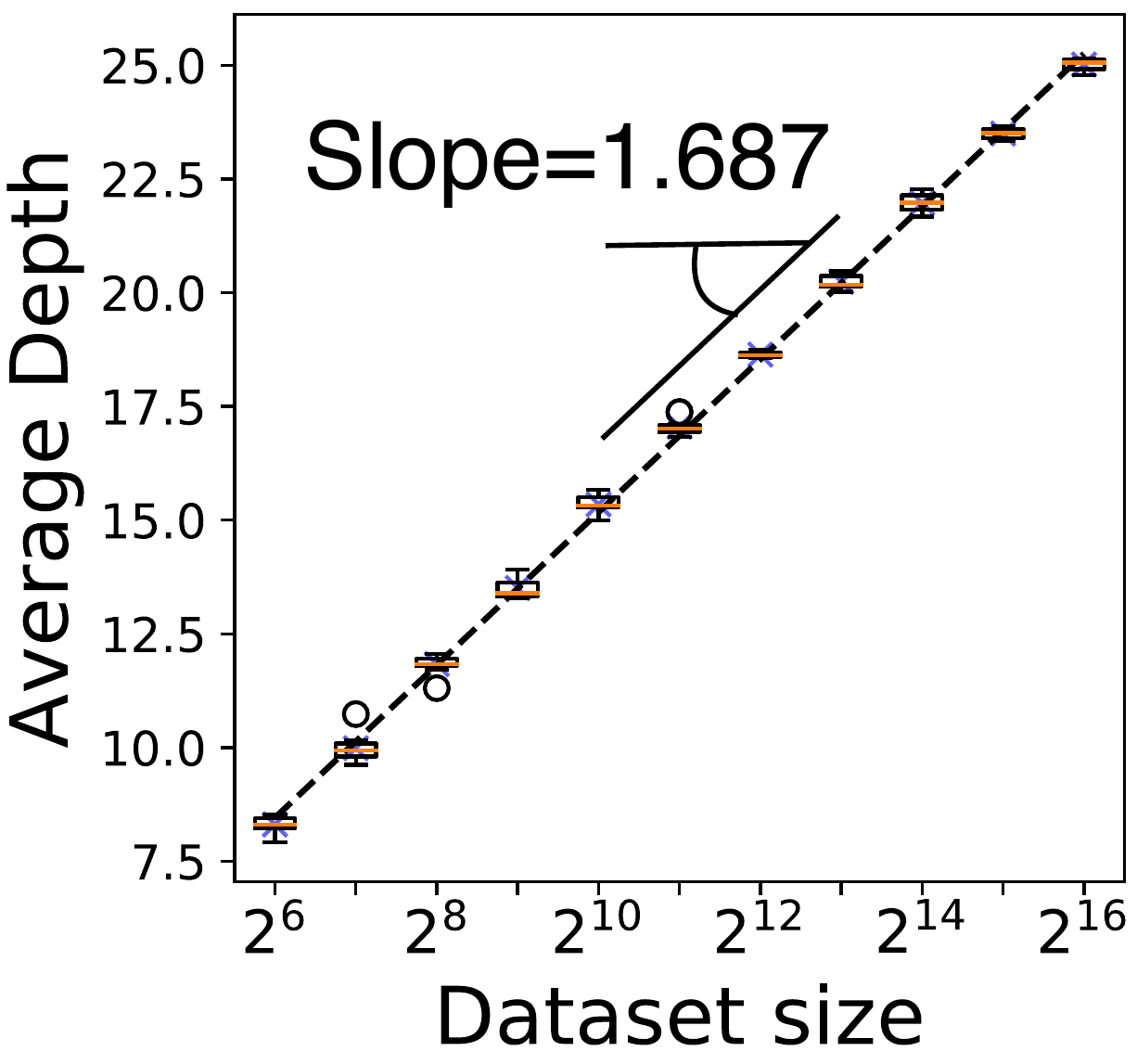}}
	\subfloat[Two Normals]
	{\label{fig:twonorm_dist}\includegraphics[scale=0.1]{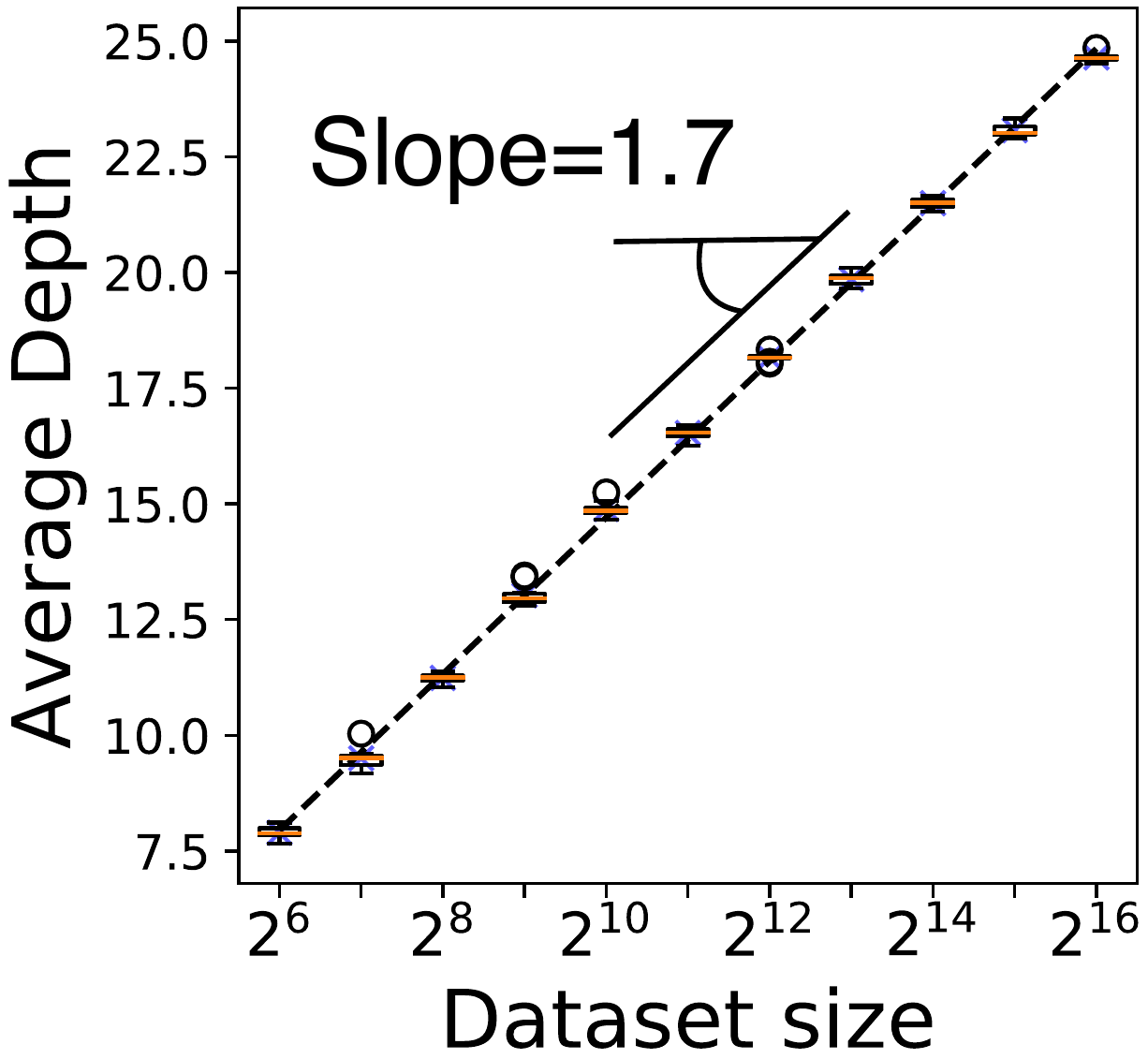}}
	\subfloat[Uniform Line]
	{\label{fig:line_dist}\includegraphics[scale=0.1]{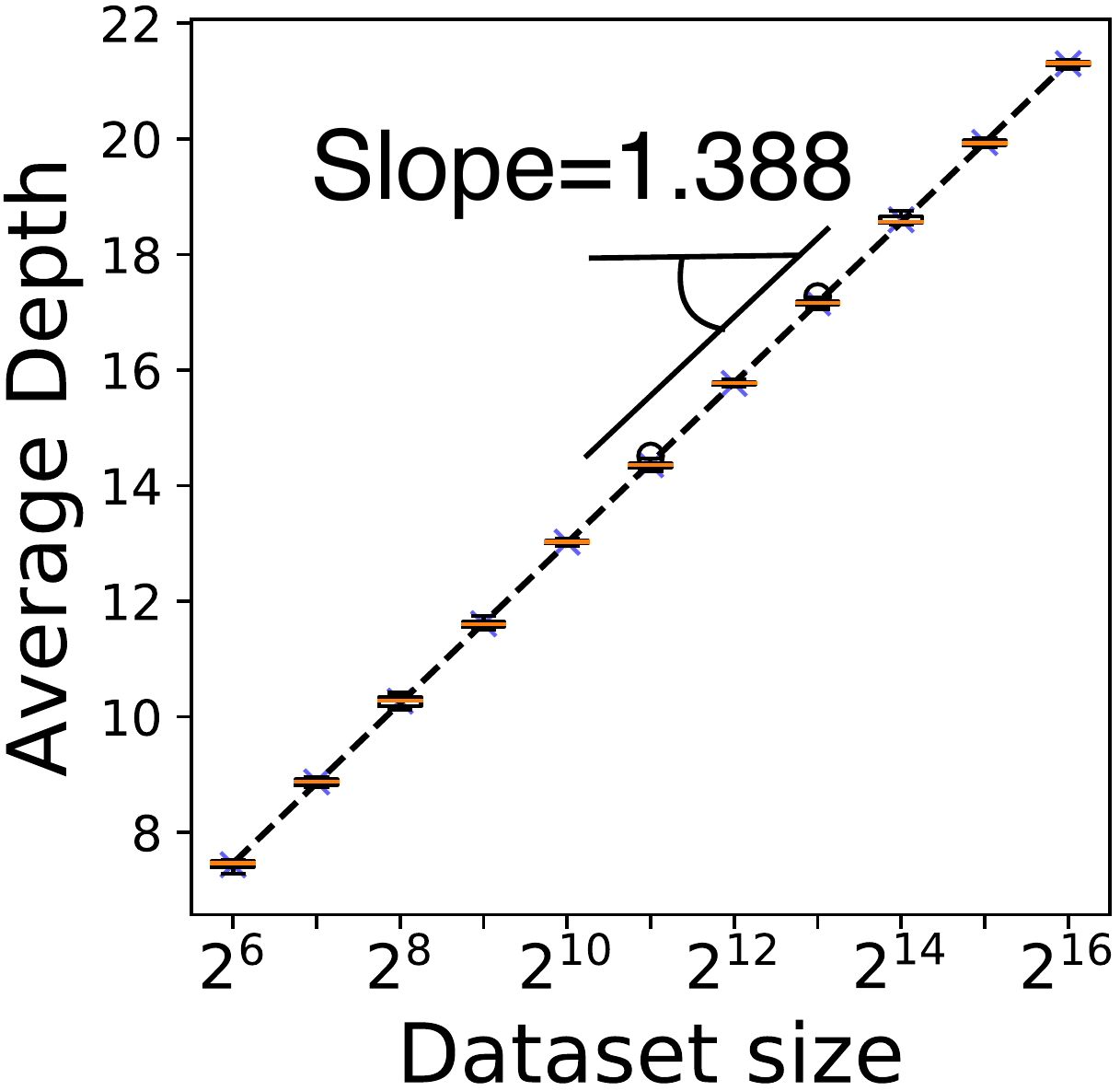}}
	\subfloat[Biased Line]
	{\label{fig:fractalline_dist}\includegraphics[scale=0.1]{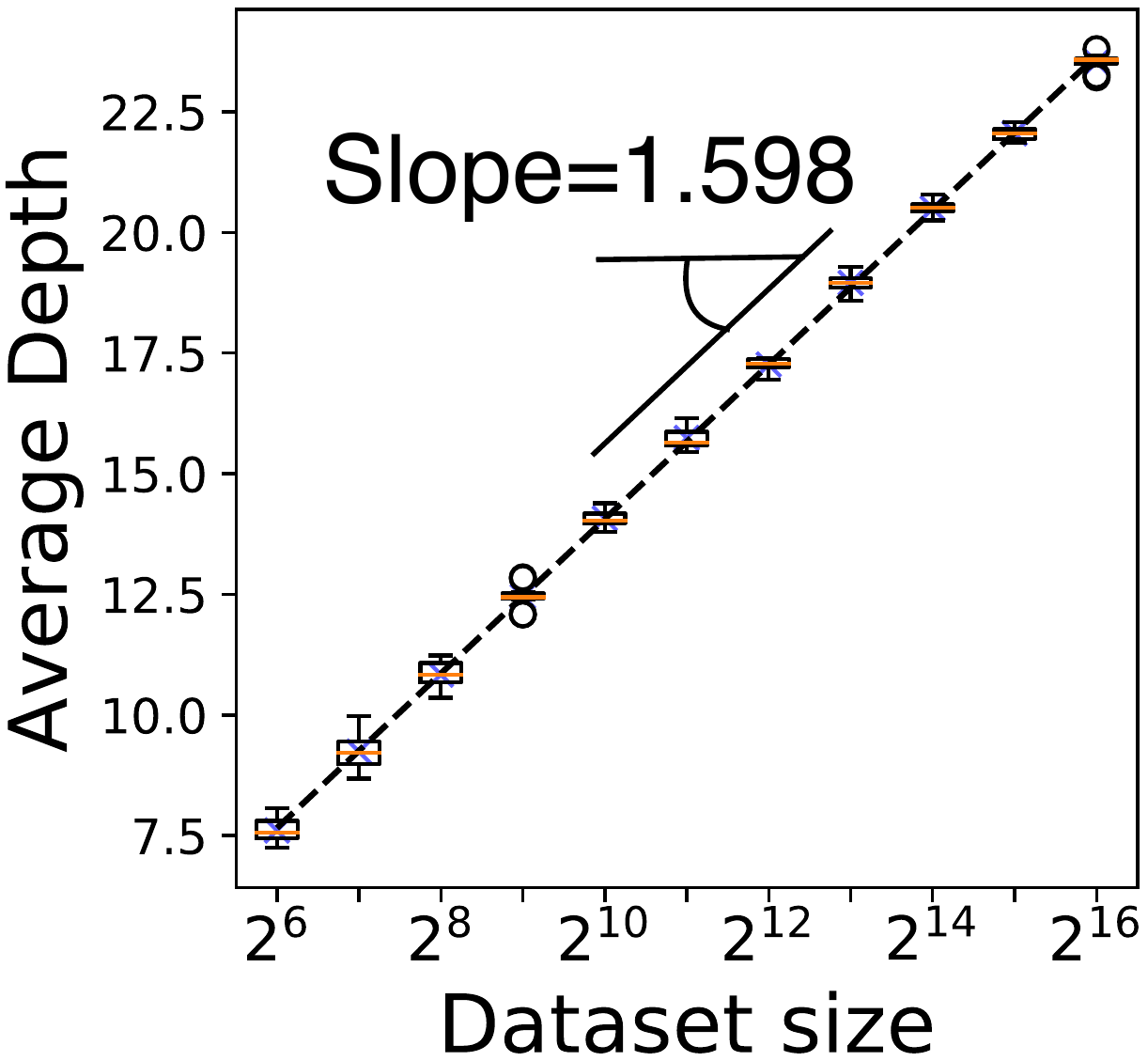}}\\
	Depth distributions\\
	\subfloat{\includegraphics[scale=0.7]{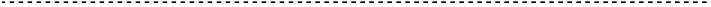}}\\
	\subfloat[Sierpinski Triangle]
	{\label{fig:triangle}\includegraphics[scale=0.17]{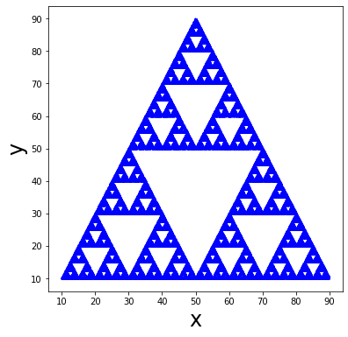}}
	\subfloat[Biased Sierpinski Triangle]
	{\label{fig:fractaltriangle}\includegraphics[scale=0.17]{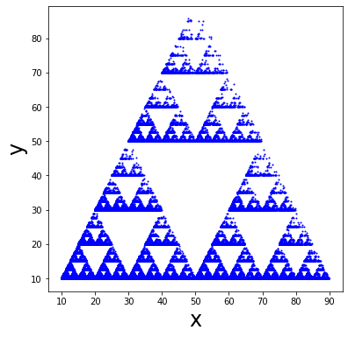}}
	\subfloat[Uniform Square]
	{\label{fig:square}\includegraphics[scale=0.17]{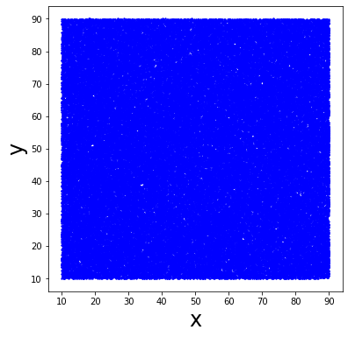}}
	\subfloat[Fern]
	{\label{fig:fern}\includegraphics[scale=0.17]{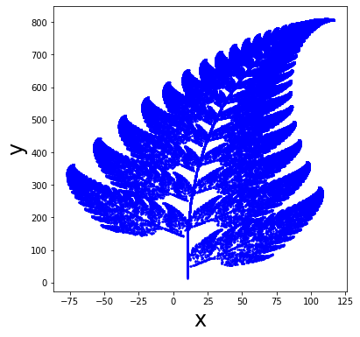}}
	\\
	\subfloat[Sierpinski Triangle]
	{\label{fig:triangle_dist}\includegraphics[scale=0.1]{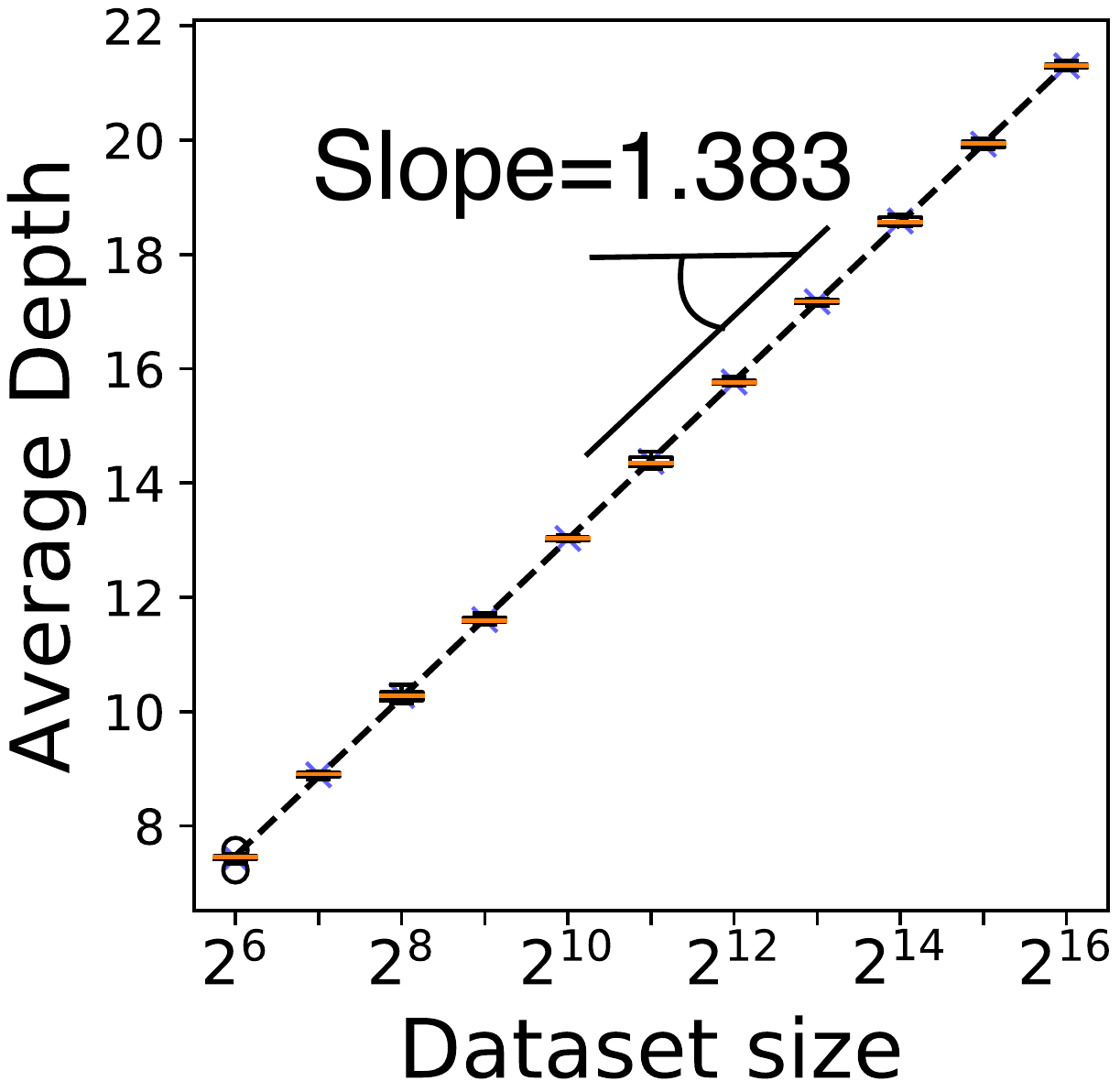}}
	\subfloat[Biased Sierpinski Triangle]
	{\label{fig:fractal_triangle_dist}\includegraphics[scale=0.1]{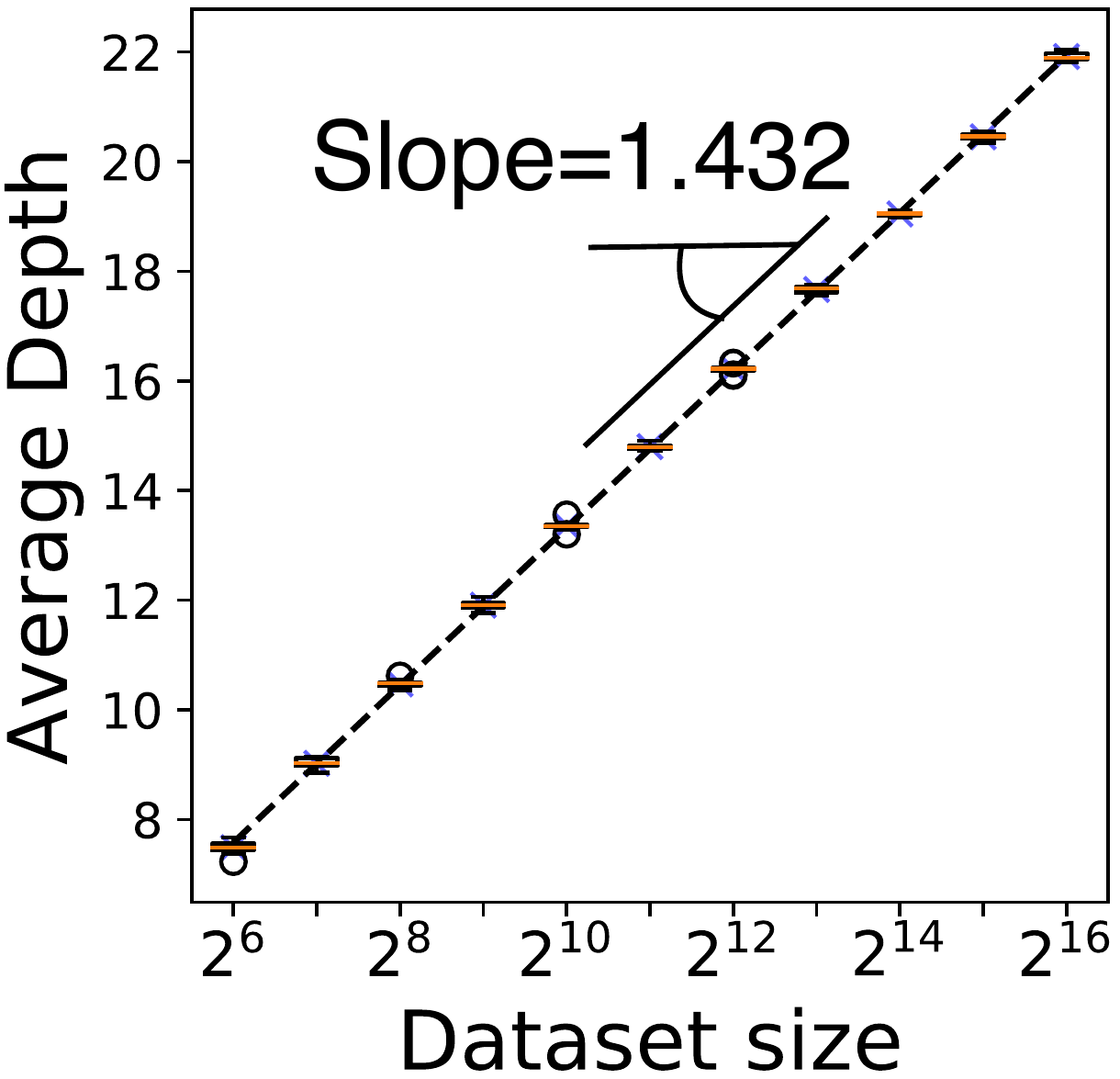}}
	\subfloat[Uniform Square]
	{\label{fig:square_dist}\includegraphics[scale=0.21]{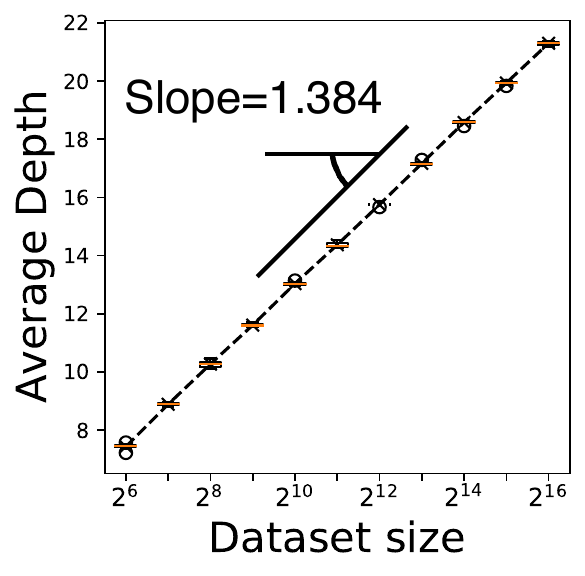}}
	\subfloat[Fern]
	{\label{fig:fern_dist}\includegraphics[scale=0.1]{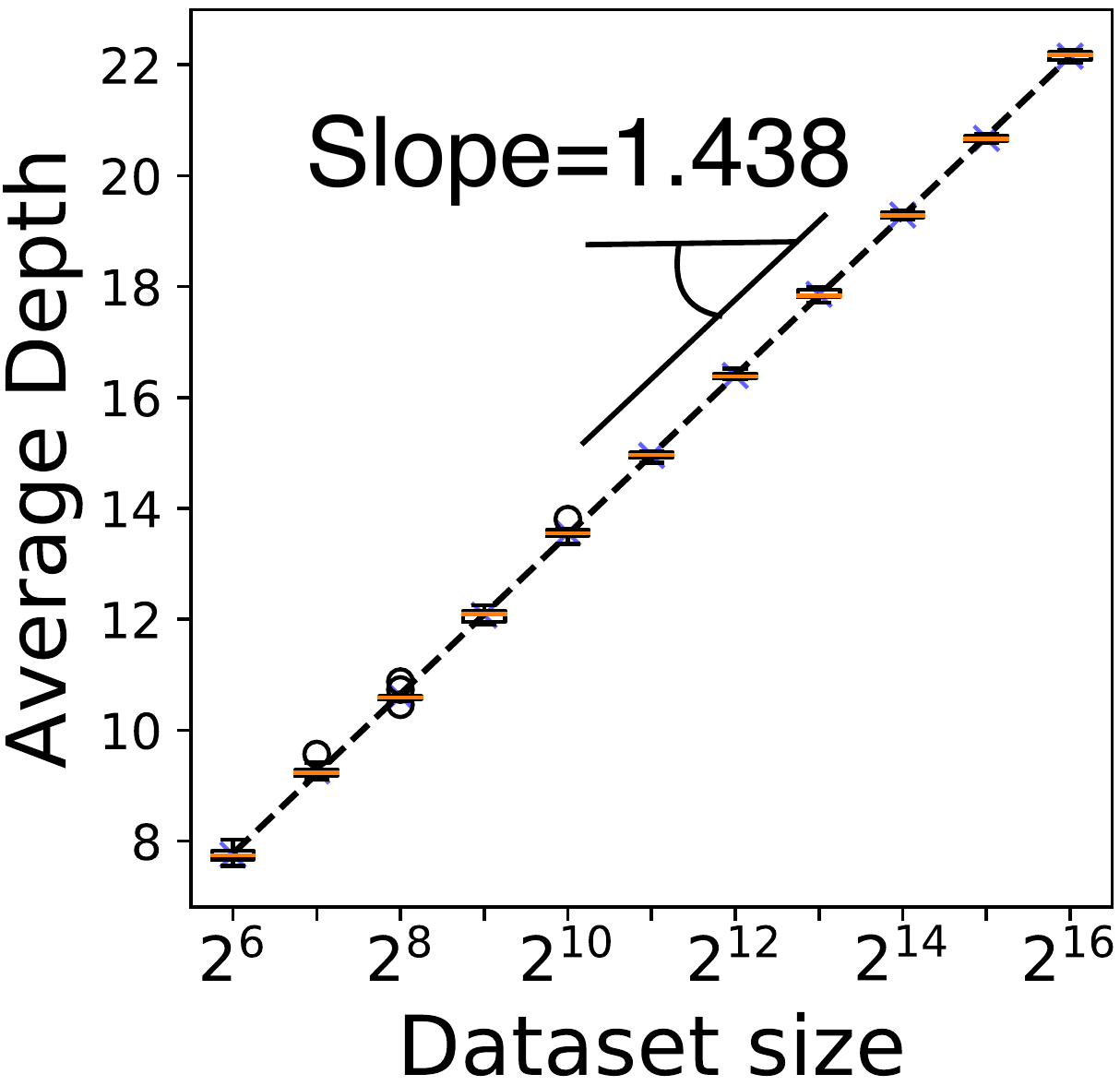}}
	\\
	Depth distributions
	\caption{Illustrating depth distribution for several diverse datasets~(including Gaussian, Uniform, multifractals).\looseness=-1}
	\label{fig:depthdist}
\end{figure}

\begin{insight}[\linearlemma~(\llshort)]
	\label{in:ltd}
    The growth of the tree depth with the logarithm counts of observations is linear irrespective of the data distribution.
\end{insight}

\noindent {\bf Justification for \llshort property:} In our attempt to explain \llshort property, we study the expected depth computation for datasets with known distributions.
However, in general, it is difficult. 
Let us consider \emph{biased} line dataset with a bias factor $b$.
Here we study a related setting: random points, but with 
fixed cuts. We refer to this model as 'fixed-cut' tree \fctree.
For this case, we can show that the PDP property holds,
and the slope grows as the `bias' factor $b$ grows.
Then, the depth of \fctree for a \emph{biased} fractal line~(data in Fig.~\ref{fig:fractalline}) obeys the following lemma.\looseness=-1
\begin{table}[]
	\centering
	\caption{ Table of symbols.\label{tab:symbols}}
	
	\resizebox{\columnwidth}{!}{
		\begin{tabular}{r l}
			\toprule
			Symbol & Definition \\
			\midrule
			$X = \{ \bs{x_i}\}$ & point cloud dataset where $\bs{x_i} \in \mathrm{R}^m$ for $i \in 1, 2, \dots n$\\
			$s(.)$ & anomaly score function for an outlier detector\\
			$h(\bs{q})$ & path length estimate for instance $\bs{q}$ as it traverses\\
			&  a depth limited \tree\\
			$\mathbb{E}[h(\bs{q})]$ & path length averaged over the ensemble\\
			$H(n)$ & depth estimation function for an \tree \\
			& containing $n$ observations\\
			\dlimit & depth limit of a \tree\\
			\bottomrule
	\end{tabular}}	
\end{table}

\begin{lemma}[Expected Depth of \fctree]
	\label{lemma:height}
	The expected tree depth $H(n, b)$ for a \emph{biased} line with a bias factor $b$ containing $n \geq 2$ data points is given as:
	\begin{align*}
		H(n, b) &= \sum_{k=0}^{n} \big[ {n \choose k} b^k (1-b)^{n-k} \times\\ &\big( \frac{k}{n} H(k, b) + \frac{n-k}{n} H(n-k, b) + 1  \big)\big]
	\end{align*}
\end{lemma}

\begin{myproof}
	Let $H(k, b)$ be the depth of \fctree with $k$ observations constructed  using $X_k \subseteq X$. 
	Since \fctree is grown via recursive partitioning on a randomly chosen attribute-value, therefore, for a biased line, $b = $~probability of a point going to left node i.e. the point less than chosen attribute-value.
	Let $k$ be the number of points partitioned onto the left node, then $n-k$ points go to right node. Define $\mathrm{B}(n, k, b) = {n \choose k} b^n (1-b)^{n-k}$ the Binomial probability for a fixed $k$. Let $f(n, k, b)$ be the estimate of the depth when $k$ observations are in left node, then $f(n, k, b) = \big( \frac{k}{n} H(k, b) + \frac{n-k}{n} H(n-k, b) + 1  \big)$ as each random partition increases depth by $1$. Therefore, the expected depth of the tree is given as $H(n, b)$ = $\sum_{k=0}^{n} f(n, k, b)\mathrm{B}(n, k, b)$. 
\end{myproof}

We denote $H(n, b) = H(n)$ for $b=1/2$. A tree with one data point would have a depth of one i.e. $H(1, b) = 1 = H(1)$; and $H(0, b) = 0 = H(0)$. In Figure~\ref{fig:biasth}, we show the effect of bias on the the (analytical) depth computed using $H(n, b)$. Notice that increase in bias -- indicating deviation from uniformity --  increases depth which matches intuition.


\begin{figure}[]
	\centering
	\includegraphics[scale=0.26]{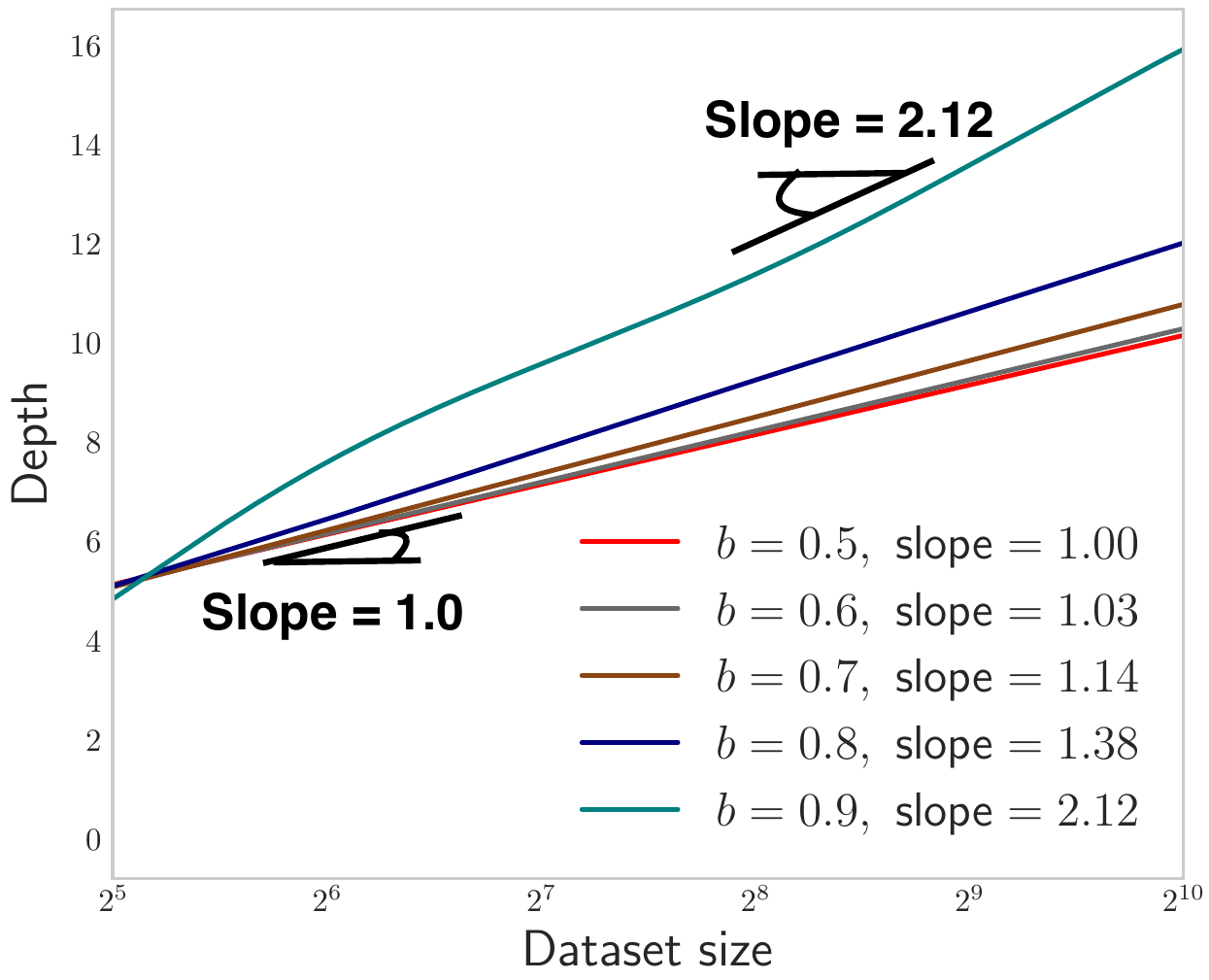}
	\caption{Depth~($\bs{H(n, b)}$) vs. Dataset size: slope increases with increase in bias for a {\em biased} line data}
	\label{fig:biasth}
\end{figure}

\begin{corollary}
	For bias $1-b$, $H(n, 1-b)$ follow the results for $H(n, b)$.
\end{corollary}


Following the \llshort{} property, the depth estimation function is given as
\begin{equation}
	\label{eq:hn}
	H(n) \approx w_0 + w_1 log_2(n)
\end{equation}
where $w_0$ and $w_1$ are parameters that we estimate for each data distribution, and $n$ is the number of instances in the dataset. 

\begin{insight}
	The slope of the linear fit varies significantly depending on the dataset distribution.
\end{insight}
For example, the slope for Uniform Line (see Fig.~\ref{fig:line_dist}) is $1.38$, while for a Uniform Square (see Fig.~\ref{fig:square_dist}) is $1.66$.
These insights lead to the following lemma.

\begin{lemma}
	\method includes \ifr as a special case.
\end{lemma}

\begin{myproof}
	In Eq.~\ref{eq:hn}, setting $w_0=  2\times 0.57 - (2(n-1)/n)$ and $w1= 2 \times log_{e}(2)$  yields the average path length function used in \ifr. Here, $0.57$ is the $\text{Euler's constant}$, and $log_{e}(2)$ accounts for the difference in log bases.
\end{myproof}

Drawing from these insights, next we present the details of our proposed anomaly detector algorithm.

\begin{table*}[!t]
    \centering
	\caption{ {\myunderline{\method wins}} as it obeys all the axioms a \genanomaly detector should follow. We compare the methods statistically, by conducting two-sample t-test based on scores obtained for points $\bs a, \bs b$. A positive difference in score indicates that the detector follows that axiom~(see Figure~\ref{fig:axiom}). \green{} indicates that the detector follows the axiom, \red{} indicates that the detector does not obey the axiom. \label{tab:axiom}}
    {\scalebox{1.1}{
    \begin{tabular}{l | c c | c c | c c || c c }
      & \multicolumn{2}{c}{\loda} & \multicolumn{2}{c}{\rcf} & \multicolumn{2}{c}{\ifr} & \multicolumn{2}{c}{\method} \\
      \hline
      & Statistic & $p$-value & Statistic & $p$-value & Statistic & $p$-value & Statistic & $p$-value \\
     \hline
        A1: Distance Axiom & \red{~~~0~~~} & \red{~~~~~~1~~~~~} & \green{~3.6~} & \green{0.002**} & \red{~~2.1~~} & \red{0.054} & \green{11.4} & \green{1.2e-9***} \\
        A2: Density Axiom & \green{7e15} & \green{2e-275***} & \red{-0.14} & \red{~~0.89~~} & \red{~~-10~~} & \red{8.6e-9***} & \green{25.2} & \green{1.7e-15***} \\
        A3: Radius Axiom & \red{~~~0~~~} & \red{~~~~~~1~~~~~}& \green{~6.4~} & \green{4.8e-6***} & \green{~11.9~} & \green{5.9e-10***} & \green{21.3} & \green{3.4e-14***} \\
        A4: Angle Axiom & \green{~~6.6~~} & \green{3.2e-6***} & \green{17.5} & \green{9.6e-13***} & \red{~-0.2~} & \red{0.83} & \green{53.7} & \green{2.5e-21***} \\
        A5: Group Axiom & \red{-14.7~} & \red{1.8e-11***} & \red{~1.1~} & \red{~~0.27~~} & \red{0.95} & \red{~~0.35~~} & \green{28.2} & \green{2.6e-16***}
    \end{tabular}}} 
\end{table*}

\section{Proposed Method }
\label{sec:meth}
For ease of exposition, we describe the algorithm in two steps -- \methodp for point anomalies, and then \method for \genanomalies.

\begin{algorithm2e}[!h]
	\SetAlgoLined
	\LinesNumbered
	\KwData{A data matrix $X$, number of \tree estimators \texttt{numTrees}, \tree depth limit \dlimit}
	\KwResult{$w_0, w_1$ of depth estimation function $H(\cdot)$ and \tree ensemble}
	Initialize $Y$ and $Z$\;
	\tcc{Estimating the function $H(\cdot)$}
	
	\For{$i = n_1, n_1 + 1, \dots$\tcp*{a small $n_1$ e.g. 10}} {
		Draw $X_{s} \subset X \text{ s.t. } |X_{s}| = 2^i$ \;
		$F_s \leftarrow$ {\sc Construct}-\tree($X_s$, $\infty$)\;
		$Z \leftarrow Z \cup~$ average depth of $F_s$ containing observations $X_s$\;
		$Y \leftarrow Y \cup~i$\;
	}
	$H(.) \leftarrow$ Fit linear regression $Y$ and $Z$\;
	$w_0, w_1 \leftarrow \texttt{coefficients}(H(.))$\;
	\tcc{Construct the ensemble of \tree}
	\For{t = 1 to \texttt{numTrees}}{
		\texttt{ensemble} $\leftarrow$  \texttt{ensemble} $\cup$ {\sc Construct}-\tree($X$, \dlimit)\;
	}
	\texttt{return } $w_0, w_1,$ \texttt{ensemble}
	\caption{\methodp-{\sc Fit} \label{algo:fit}}
\end{algorithm2e}

\subsection{Point anomalies -- \methodp }
\label{subsec:algo}
Given the observations $X = \{ \bs{x_1}, \dots, \bs{x_M} \}$ where $\bs{x_i} \in \mathrm{R}^m$, \methodp's goal is to detect and assign anomaly score to outlier points. \methodp uses an ensemble of depth-limited randomized tree \tree ({\S\ref{subsec:insights}}) that recursively partition instances in $X$. 

\begin{definition}[Depth Limited \tree]
	An \tree that is constructed by recursively partitioning the given set of observations $X$ until a depth limit \dlimit{} is reached or the {\em leaf} nodes contain exactly one instance.
\end{definition}
As evidenced in prior works, the random trees induce shorter path lengths (number of steps from root node to leaf node while traversing the tree) for anomalous observations since the instances that deviate from other observations are likely to be partitioned early. Therefore, a shorter average path length from the ensemble would likely indicate an anomalous observation. Anomaly detection is essentially a ranking task where the rank of an instance indicates its relative degree of anomalousness. We next design anomaly score function for our algorithm to facilitate ranking of observations.\\
\begin{algorithm2e}[!h]
	\SetAlgoLined
	\LinesNumbered
	\KwData{A data matrix $X$,\dlimit, \texttt{currDepth:0}}
	\KwResult{\tree}
	Initialize \tree\;
	\uIf{\dlimit{} $\leq$ \texttt{currDepth} \text{or} $|X| \leq1$}{
			\texttt{return } a leaf node of size $|X|$ 
	}
	\Else{pick an attribute at random from $X$\;
		pick an attribute value at random\;
		$X_l\leftarrow$ set of points on the left ($<$) of the chosen attribute-value pair\;
		$X_r\leftarrow$ set of points on the right ($\geq$) of the chosen attribute-value pair\;
		\texttt{left}  $\leftarrow$ {\sc Construct}-\tree($X_l$,\dlimit, \texttt{currDepth + 1})\;
		\texttt{right} $\leftarrow$ {\sc Construct}-\tree($X_r$,\dlimit, \texttt{currDepth + 1})
		
			\texttt{return } an internal node with \{\texttt{left}, \texttt{right}, \{chosen attribute-value pair\}\}
		
	}
	\caption{{\sc Construct}-\tree \label{algo:tree}}
\end{algorithm2e}
\noindent {\bfseries Proposed Anomaly Score.~}
We construct anomaly score using the path length $h(\bs{q})$ for each instance $\bs{q} \in \mathrm{R}^m$ as it traverses a depth limited \tree.  The path length for $\bs{q}$ is $h(\bs{q}) = h_{0} + H(l_{busy}) \text{ if } l_{busy} > 1; \text{ otherwise } h(\bs{q}) = h_{0}$ where  $h_{0} $ is
the number of edges $\bs{q}$ traverses from \emph{root} node to \emph{leaf} node that contains $l_{busy}$ points in a depth limited \tree. When $l_{busy} > 1$, we estimate the expected depth from the leaf node using $H(l_{busy})$ (uses Eq.~\ref{eq:hn}). We normalize $h(\bs{q})$ by the average tree height $H(n)$ (height of \tree containing $n$ observations) for the depth limited \tree ensemble to produce an anomaly score $s(\bs{q}, n)$ for a given observation $\bs{q}$. Referring to the \llshort insights we presented in Section \S\ref{subsec:insights}, we estimate the data dependent $H(\cdot)$ using Eq.~\ref{eq:hn} since the tree depth grows linearly with the number of observations (in $log_2$) in the tree (see Figure~\ref{fig:depthdist}). The slope of the linearity is characterized by underlying data distribution; each distribution follows a linear growth. The score function is
\begin{equation}
    s(\bs{q}, n) = 2^{- \frac{\mathrm{E}[h(\bs{q})]}{H(n)}}
\end{equation}
where $\mathrm{E}[h(\bs{q})]$ is the average path length of observation $\bs{q}$ in the \tree ensemble, $n$ is number of data points used to construct each \tree, and $H(n)$ is the function for estimating depth of the tree given in Eq.~\ref{eq:hn}.\\

\noindent {\bfseries \methodp Parameter Fitting.~} \method is a depth limited \tree ensemble.
The algorithm for fitting \methodp parameters is provided in Algorithms~\ref{algo:fit} and \ref{algo:tree}.
\begin{algorithm2e}
	\caption{\methodp-Scoring \label{algo:predict}}
\SetAlgoLined
\LinesNumbered
\KwData{A data matrix $X$, \tree ensemble}
\KwResult{Anomaly scores \texttt{scores} for observations in $X$}
Initialize \texttt{depths}\;
Initialize \texttt{scores}\;
Initialize \leaf\;
$n \leftarrow$ \texttt{numSamplesIn\tree}\;
\For{$\bs{x} \in X$}{
    \texttt{depths} $\leftarrow$ \texttt{depths} $\cup$ compute path-lengths for $\bs{x}$ (see \S\ref{subsec:algo})\;
    
    \leaf $\leftarrow$ \leaf$\cup$ compute number of samples in \texttt{leaf} where traversal of $\bs{x}$ terminated \;
}
\For{depth $\in$ \texttt{depths}, $l \in$ \leaf}{
    $h = $depth$ + H(l)$\;
    $s = 2^{\frac{-h}{H(n)}}$\;
    \texttt{scores} $\leftarrow$ \texttt{scores} $\cup~ s$\;
}
	\texttt{return } \texttt{scores}\;

\end{algorithm2e}

\noindent {\bfseries \methodp Scoring.~} To assign anomaly scores to the instances in a data matrix $X$, the expected path length $\mathrm{E}(h(\bs{q}))$ for each instance $\bs{q} \in X$. $\mathrm{E}(h(\bs{q}))$ is estimated by averaging the path length after tree traversal through each \tree in \method ensemble. We outline the steps to assign anomaly score to a data point using \methodp in Algorithm~\ref{algo:predict}.

\begin{algorithm2e}[]
	\caption{\method \label{algo:g2o}}
	\SetAlgoLined
	\LinesNumbered
	Initialize $n \leftarrow |X|$\;
	
	\tcc{Step 0: Fit a sequence of \methodp}
	\For{$qr \in \{1, 1/2, 1/4, \cdots\}$}{
		Draw $X_s \subset X$ s.t. $|X_s| = n \times qr$\;
		\methodp-$ensembles$ $\leftarrow$ \methodp-$ensembles$ $\cup$ \methodp-\sc{Fit} ($X_s$, ., .)\;
	}

	\tcc{Step 1: create \xray plot}

	\For{e $\in$ \methodp-$ensembles$}{
		\tcc{generate score for specific qualification rate}
		\texttt{scores} $\leftarrow$ \texttt{scores} $\cup$ \methodp-Scoring(X, e)\;
	}
	\tcc{Step 2: Apex extraction}
	\tcc{max score and qualification rate for each point across qualified datasets}
	\texttt{max\_scores, max\_qr}  $\leftarrow\arg\max(	\texttt{scores})$ \;
	
	\tcc{select points with max score above threshold}
	\texttt{candidate-points} $\leftarrow X$[\texttt{max\_scores} $\geq$ $threshold$];
	
	\tcc{Step 3: Outlier grouping}
	\For{r $\in$ unique(\texttt{max\_qr})}{
		\texttt{candidate-points\_r} \tcp*{candidate points at this qualification rate}
		\tcc{identify more than one \group per qualification rate}
		\texttt{clusters} $\leftarrow$ $cluster$ \texttt{candidate-points\_r}\;
	}
	\tcc{Step 4: Compute iso-curves}
	\For{$cl$ $\in$ \em{\texttt{clusters}}}{
		\tcc{points similar to outlier at (score=1, qr=1) is more anomalous }
		\texttt{iso\_scores} $\leftarrow$ $\frac{2 - ManhattanDistance([\frac{\log_2{max\_qr(\bs{a})}}{10}+1, max\_score(\bs{a})], [1, 1])}{2}$  $\forall \bs{a} \in cl $\;
	}
	\tcc{Step 5: Scoring}
	assign $ \texttt{scores}  \leftarrow median(\texttt{iso\_scores}(cl)) \;\;\forall cl \in \texttt{clusters}\  $
	
\end{algorithm2e}

\subsection{Full algorithm -- \method }
\methodp can spot point-anomalies.
How can design an algorithm that can spot
both point- as well as group-anomalies,
simultaneously?

The main insight is to exploit the less-appreciated
ability of sampling to drop outliers, with high probability.
How can we use this property to spot group-anomalies,
of size, say $n_g$ (in a population of $n$ data points)?
The idea is that, with a sampling rate of $n_g/n$,
a point $\bs{a}$ of the \group
will probably be stripped of its cohorts,
and thus behave like a point-anomaly, exhibiting
a high anomaly score.
For dis-ambiguation versus the sampling of 
\methodp, we will refer to this sampling process
as `{\em qualification}', and to the corresponding rate
as $qr$= {\em qualification rate}.

In more detail, to determine whether point $\bs{a}$ belongs
to a group-anomaly, we compute its (\methodp) score $s(\bs{a}, qr)$
for several qualification rates $qr$; when the score peaks
(say, at rate $n_g/n$) then $n_g$ is roughly the size
of the group-anomaly (= micro-cluster) that $\bs{a}$ belongs to. Some definitions:
\begin{definition}[\xray -line]
For a given data point $\bs{a}$, 
the \xray line is the function (score($\bs{a}$, qr) vs qr).
\end{definition}
\begin{definition}[\xray plot]
	For a cloud of $n$ points, the \xray plot is the 2-d plot
	of all the $n$ \xray-lines (one for each data point)
\end{definition}See Figure~\ref{fig:syn:xray} for an example.
\begin{definition}[\sxray]
{Apex} of point $\bs{a}$ is the point (score, qr) with the highest
anomaly score.
\end{definition}See Figure~\ref{fig:syn:sxray} for an example.



Algorithm ~\ref{algo:g2o} describes the steps of the proposed \method.
In summary, we find the \xray plot (Step 1) and then find the apex point for every data point $\bs{a}$
(Step 2);
keep the ones with high apex
and then cluster the corresponding data points
(Step 3); and then assign scores to the each \group (Step 4 and Step 5).

Figure~\ref{fig:syntheticgrp} illustrates the steps in \method on a synthetic dataset that has two anomalous \group{s} along with several point anomalies. \begin{figure*}[t]
	\centering
	\subfloat[Synthetic data heatmap]
	{\label{fig:syn:hex}\includegraphics[width=1.17in]{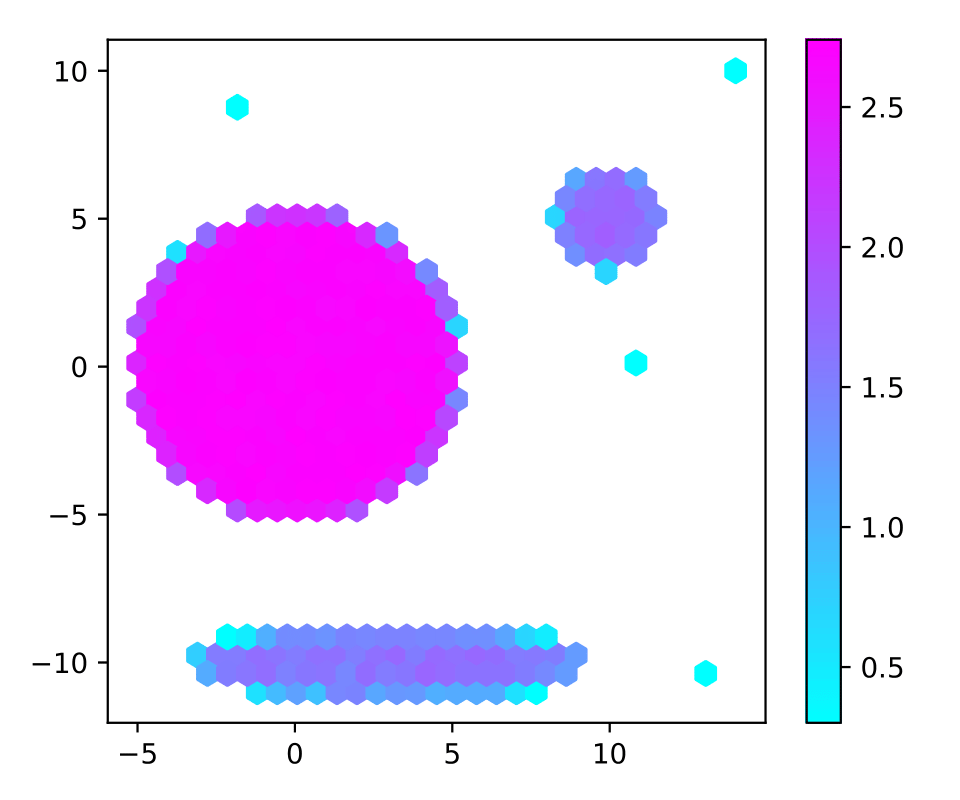}}
	\subfloat[Step 1: \xray plot]
	{\label{fig:syn:xray}\includegraphics[scale=0.09]{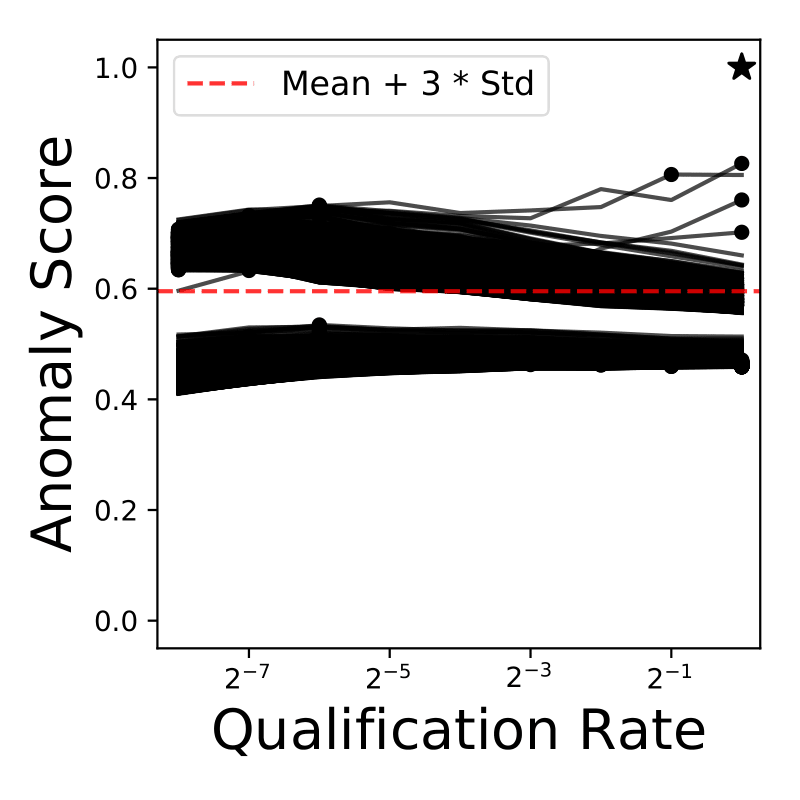}}
	\subfloat[Step 2: Apex extraction]
	{\label{fig:syn:sxray}\includegraphics[scale=0.09]{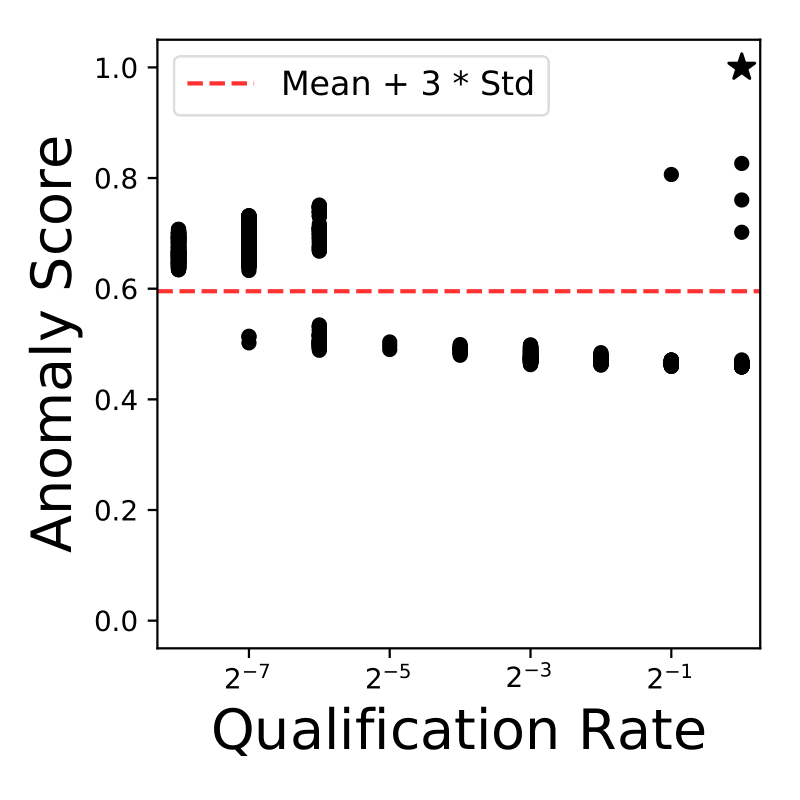}}
	\subfloat[Step 3: Outlier grouping]
	{\label{fig:syn:post}\includegraphics[width=1.in]{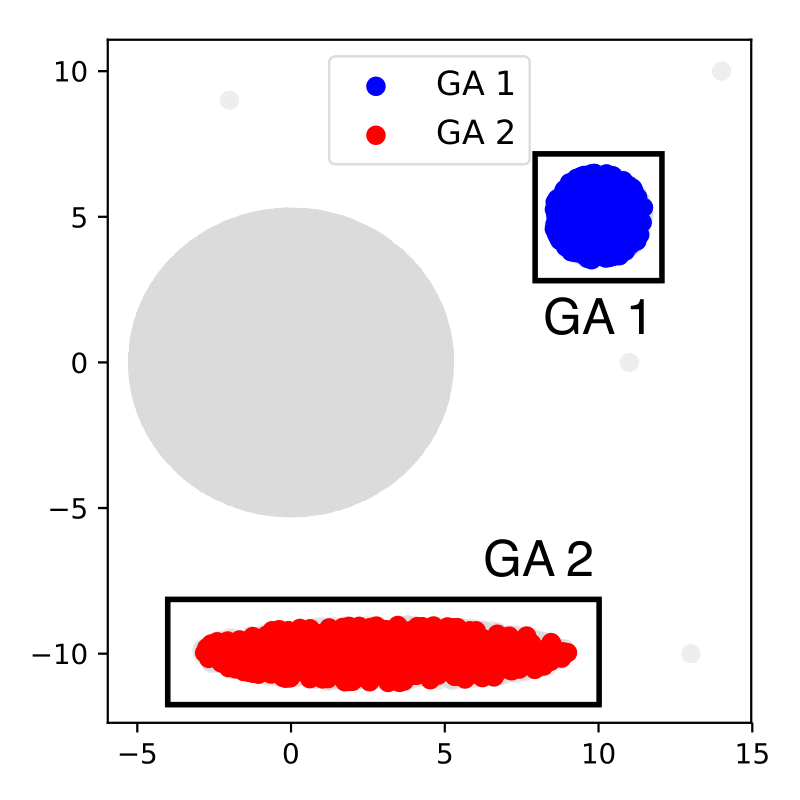}}
	\subfloat[Step 4: Anomaly iso-curves]
	{\label{fig:syn:iso}\includegraphics[scale=0.09]{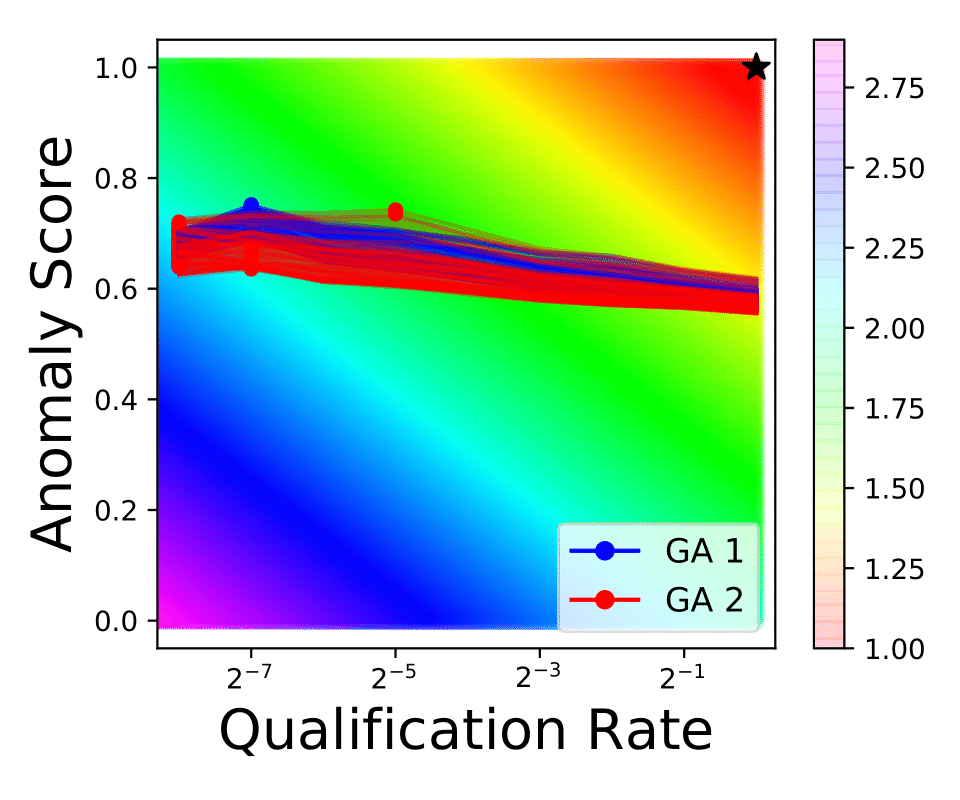}}
	\subfloat[Step 5: Scoring]
	{\label{fig:syn:gs}\includegraphics[scale=0.09]{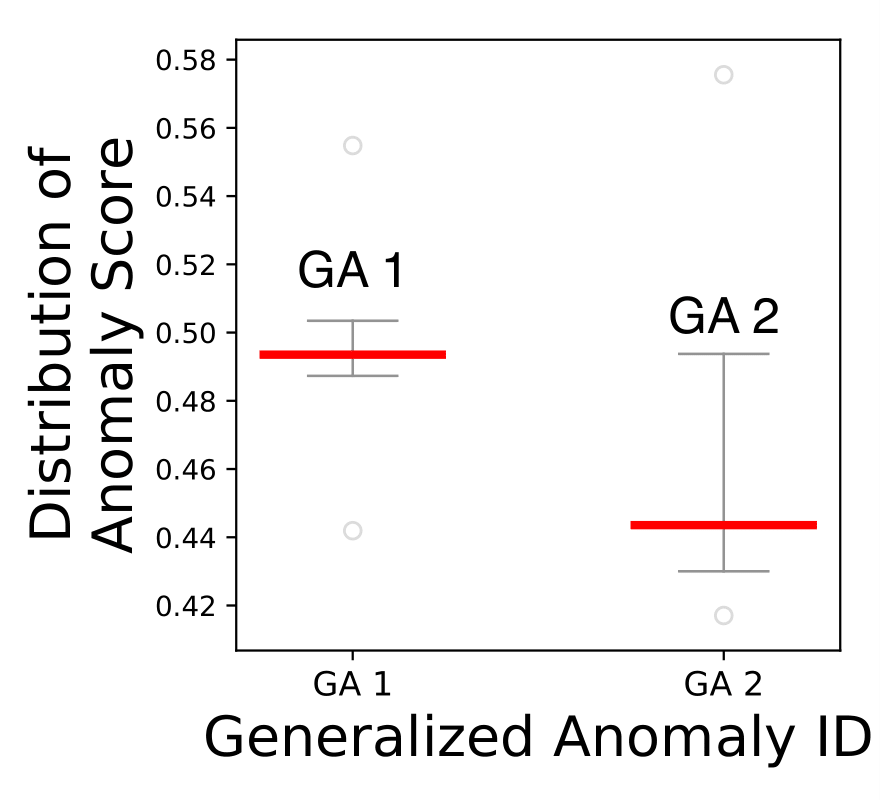}}
	\caption{\myunderline{\method works.} Illustration of \method on synthetic dataset\label{fig:syntheticgrp}}
\end{figure*}

Figure~\ref{fig:syn:xray} finds the \xray plot and Figure~\ref{fig:syn:sxray} shows the apex with the red threshold line. We find two \group{s} after applying clustering (dbscan~\cite{schubert2017dbscan} in our implementation) shown in color red, and blue in Figure~\ref{fig:syn:post}. Then we compute the similarity of points in \xray plot representation in each cluster to the theoretically most anomalous point at score$=1$, qr$=1$ (see iso curves in Figure~\ref{fig:syn:iso}), and then assign \genanomaly score using the median of the similarity scores as shown in Figure~\ref{fig:syn:gs}. \method correctly assigns higher score to GA1 (blue cluster in Figure~\ref{fig:syn:gs}) which contains $1000$ points as compared to GA2 (red cluster in Figure~\ref{fig:syn:gs}) containing $2000$ points (also see Axiom A5). For ease of visualization, we do not show \ptanomalies in this plot.

\section{Experiments} 
\label{sec:exp}
\begin{table}[]
	\caption{Benchmark datasets summary.\label{tab:bmdatatable}}
	\centering{\resizebox{0.8\columnwidth}{!}{
			\begin{tabular}{l | r r r }
				\hline
				Datasets & \#Samples & Dimension & \% Outliers \\
				\hline \hline
				\multicolumn{4}{c}{$\text{Size} < 3000$} \\
				\hline
				arrhythmia  & 452   & 274 & 14.6\% \\
				cardio      & 1831  & 21  & 9.6\% \\
				glass       & 214   & 9   & 4.2\% \\
				ionosphere  & 351   & 33  & 35.9\% \\
				letter      & 1600  & 32  & 6.3\% \\
				lympho      & 148   & 18  & 4.1\% \\
				pima        & 768   & 8   & 34.9\% \\
				vertebral   & 240   & 6   & 12.5\% \\
				vowels      & 1456  & 12  & 3.4\% \\
				wbc         & 378   & 30  & 5.6\% \\
				breastw     & 683   & 9   & 35\% \\
				wine        & 129   & 13  & 7.8\% \\
				\hline
				\multicolumn{4}{c}{$\text{Size} \geq 3000$} \\
				\hline
				mnist       & 7603  & 100 & 9.2\% \\
				musk        & 3062  & 166 & 3.2\% \\
				optdigits   & 5216  & 64  & 2.9\% \\
				pendigits   & 6870  & 16  & 2.3\% \\
				satellite   & 6435  & 36  & 31.6\% \\
				satimage-2  & 5803  & 36  & 1.2\% \\
				shuttle     & 49097 & 9   & 7.2\% \\
				annthyroid  & 7200  & 6   & 7.4\% \\
				cover       & 286048 & 10   & 0.96\% \\
				http        & 567498 & 3    & 0.39\% \\
				mammography & 11183 & 6   & 2.3\% \\
				smtp        & 95156 & 3   & 0.032\% \\
				speech      & 3686  & 400 & 1.7\% \\
				thyroid     & 3772  & 6   & 2.5\% \\
				\hline
			\end{tabular}
	}}
\end{table}

We evaluate our method through extensive experiments on a set of datasets from real world use-cases. 
We now provide dataset details and the experimental setup, followed by the experimental results.
\subsection{Dataset Description}
\noindent\textbullet $\;${\bfseries Epilepsy Dataset.~} We analyzed intracranial electroencephalographic (EEG) signals recorded at the Epilepsy Monitoring Unit of a large public university
from one patient with refractory epilepsy.  Electrodes were stereotactically placed in the brain and EEG signals were then recorded across 122 electrode contacts at a sampling rate of 2KHz with focal region in the right temporal lobe.

\textbullet $\;${\bfseries Benchmark Datasets.~} Our benchmark set consist of \numdatasets{} real-world outlier detection datasets from ODDS repository~\cite{Rayana:2016}. The datasets cover diverse application domains and have diverse range dimensionality and outlier percentage~(summarized in Table~\ref{tab:bmdatatable}). The ODDS datasets provide ground truth outliers that we use for the quantitative evaluation of the methods.

\begin{figure}[t]
	\centering
	\subfloat[Dataset size $< 3000$.]
	{\label{}
		\includegraphics[scale=0.38]{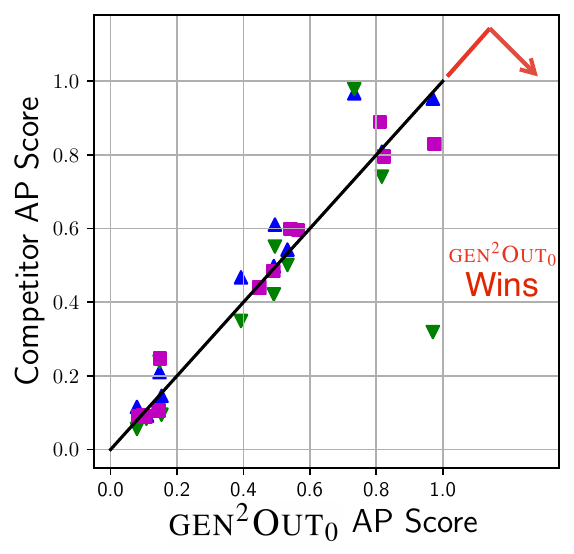}
		\includegraphics[scale=0.38]{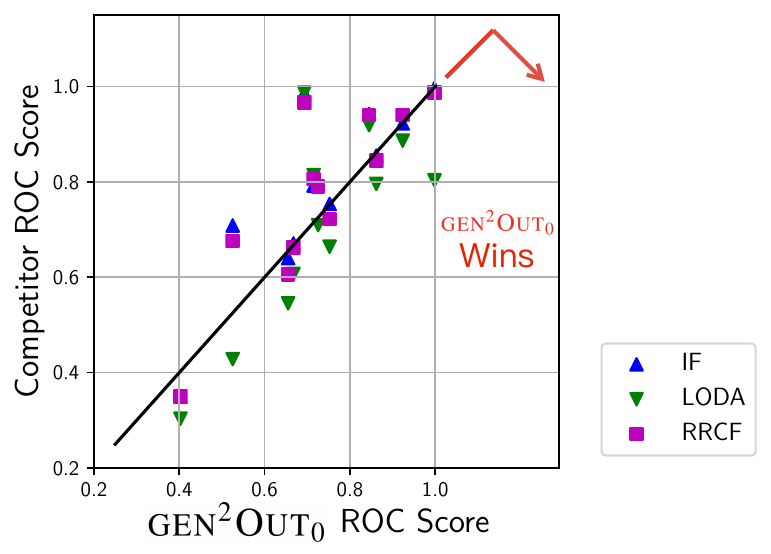}}\\
	\subfloat[Dataset size $\geq 3000$ where \rcf doesn't scale.]
	{\label{}
		\includegraphics[scale=0.38]{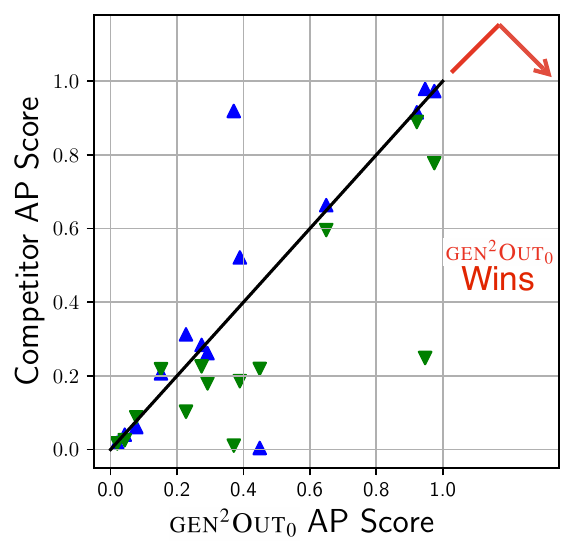}
		\includegraphics[scale=0.38]{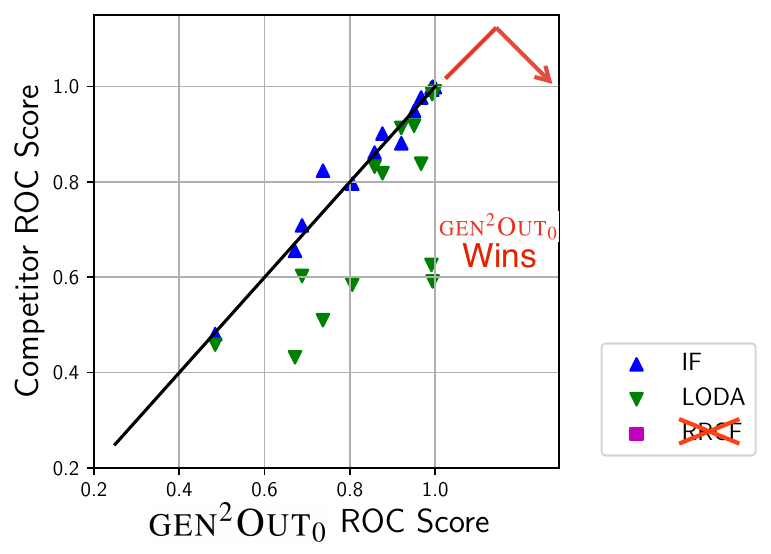}}
	\caption{{\myunderline{\methodp wins}}. We plot average precision (AP) and area under the ROC curve for \methodp against the same metric of the competitors (none of which obey all our axioms). Points representing benchmark datasets are below the line for the majority of datasets.}
	\label{fig:aproc}
\end{figure}

\begin{figure*}
	\centering
	\subfloat[Data heatmap]
	{\label{fig:http:hex}\includegraphics[scale=0.09]{NEW_FIG/http_hexbin.png}}
	\subfloat[\xray plot]
	{\label{fig:http:xray}\includegraphics[scale=0.09]{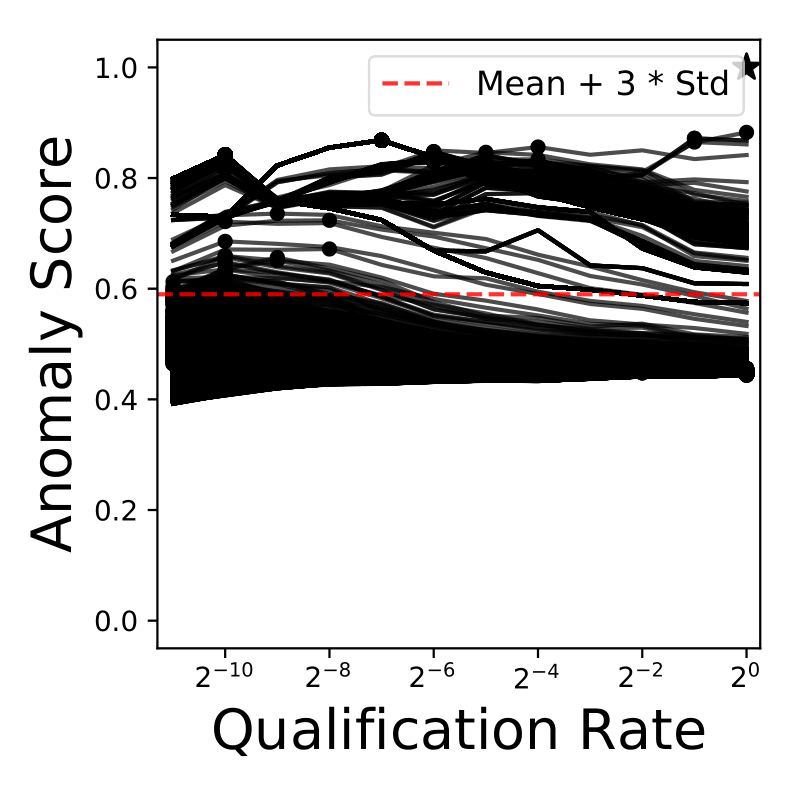}}
	\subfloat[Apex extraction]
	{\label{fig:http:sxray}\includegraphics[scale=0.09]{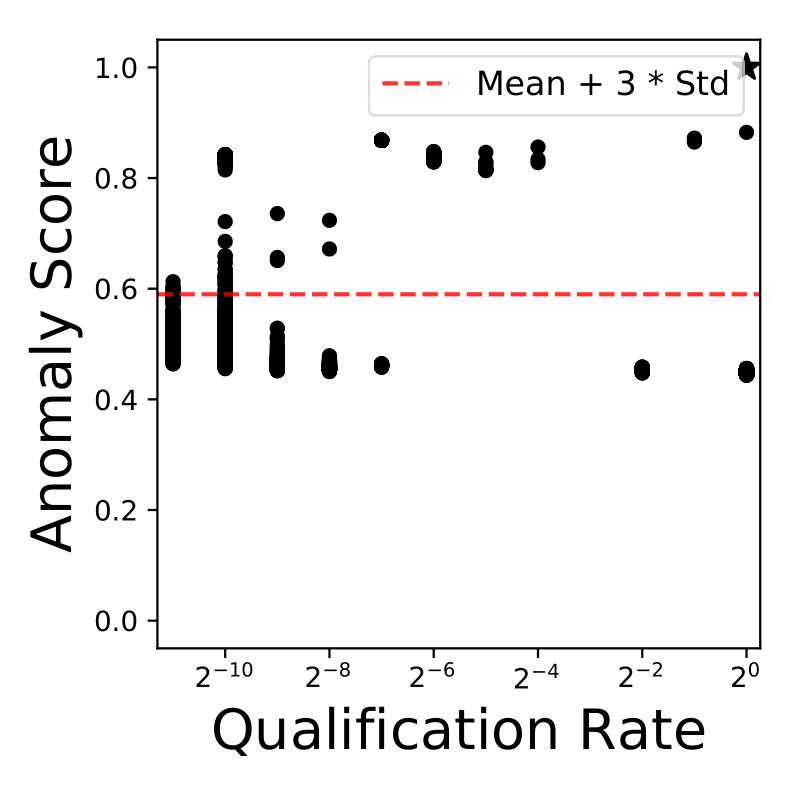}}
	\subfloat[Outlier grouping]
	{\label{fig:http:post}\includegraphics[scale=0.042]{NEW_FIG/http_dbscan_copy.png}}
	\subfloat[Anomaly iso-curves]
	{\label{}\includegraphics[scale=0.09]{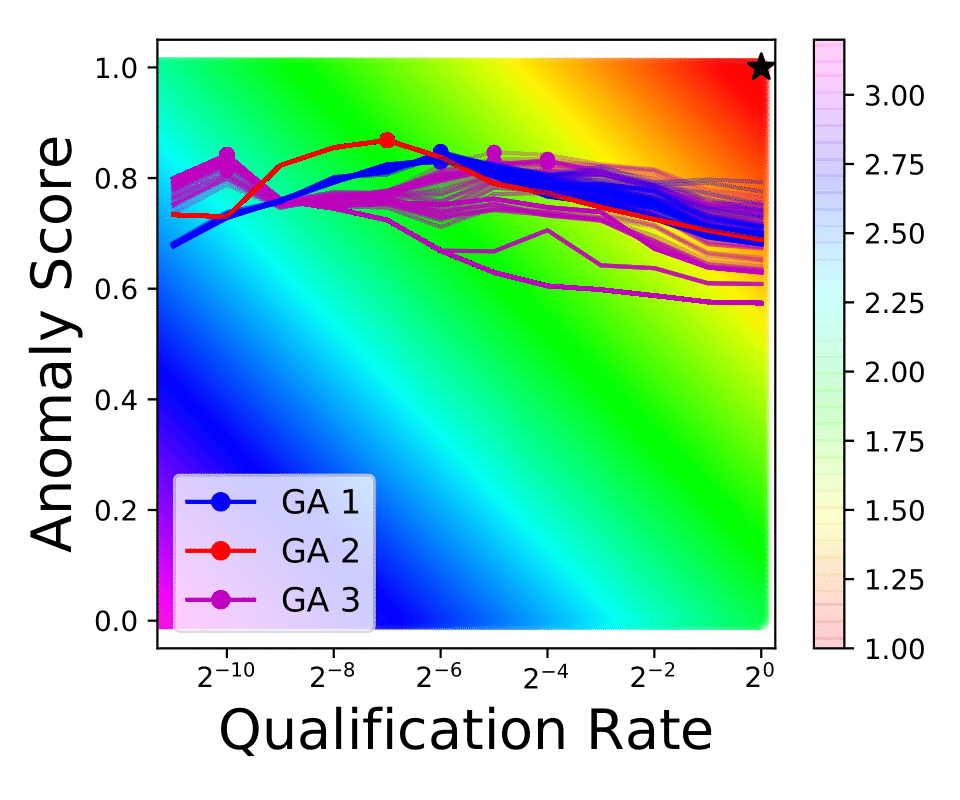}}
	\subfloat[Scoring]
	{\label{fig:http:gs}\includegraphics[scale=0.09]{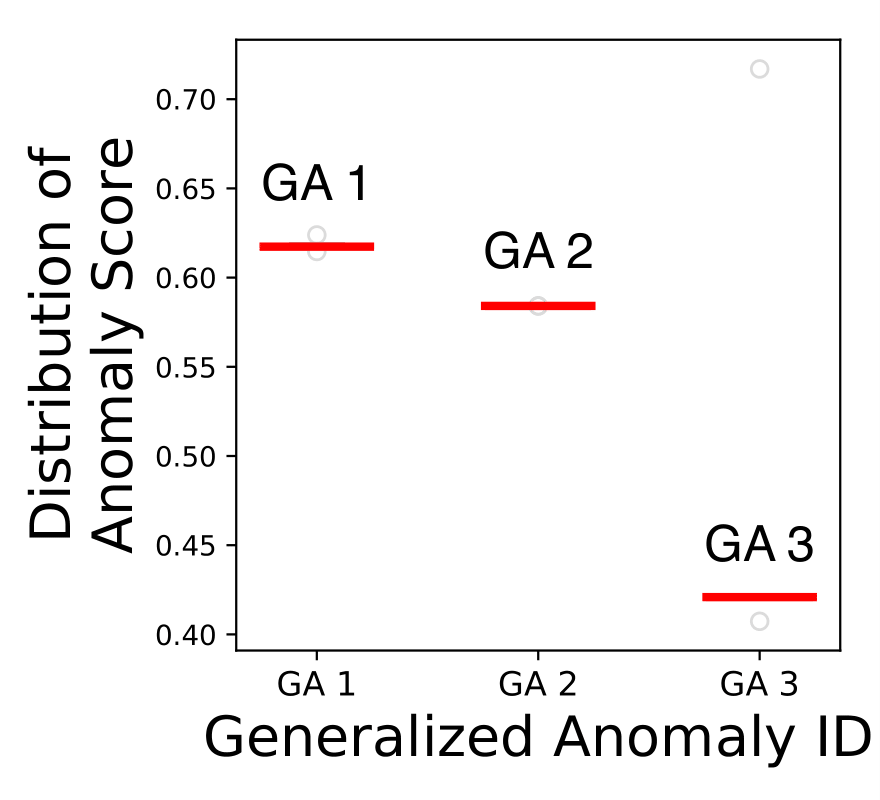}}
	\caption{\myunderline{\method detects DDoS attacks} on intrusion detection \http dataset}
\end{figure*}
\begin{figure*}
	\centering
	\subfloat[Heatmap of tSNE representation of data]
	{\label{fig:eeg:heat}\includegraphics[scale=0.09]{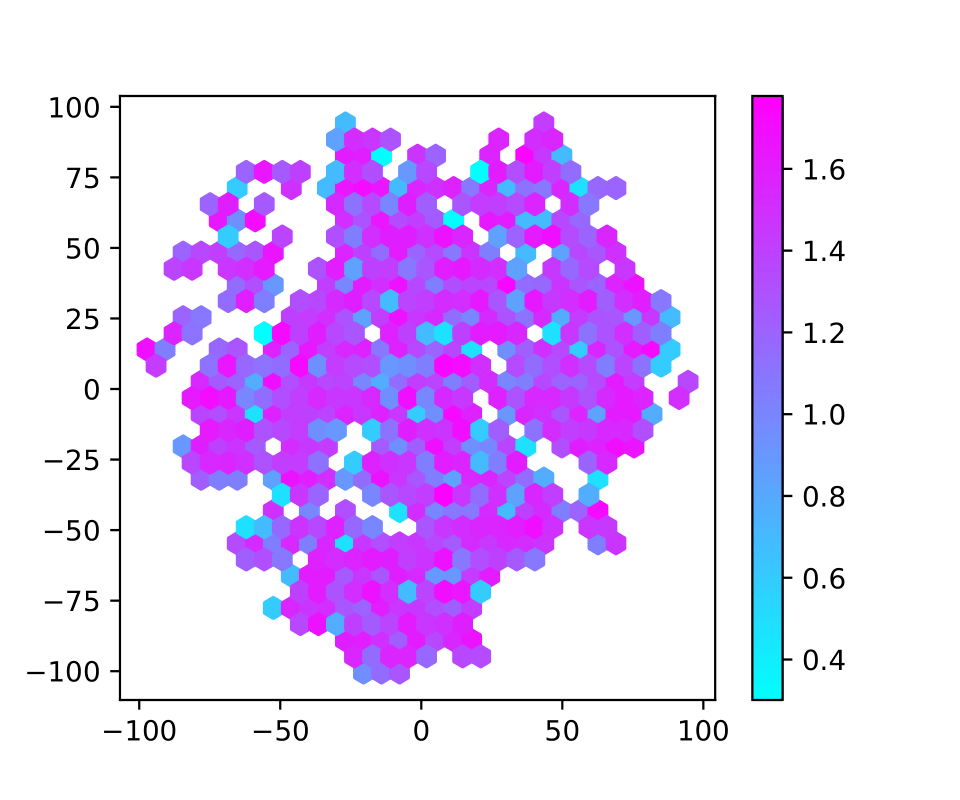}}
	\subfloat[\xray plot]
	{\label{fig:eeg:xray}\includegraphics[scale=0.09]{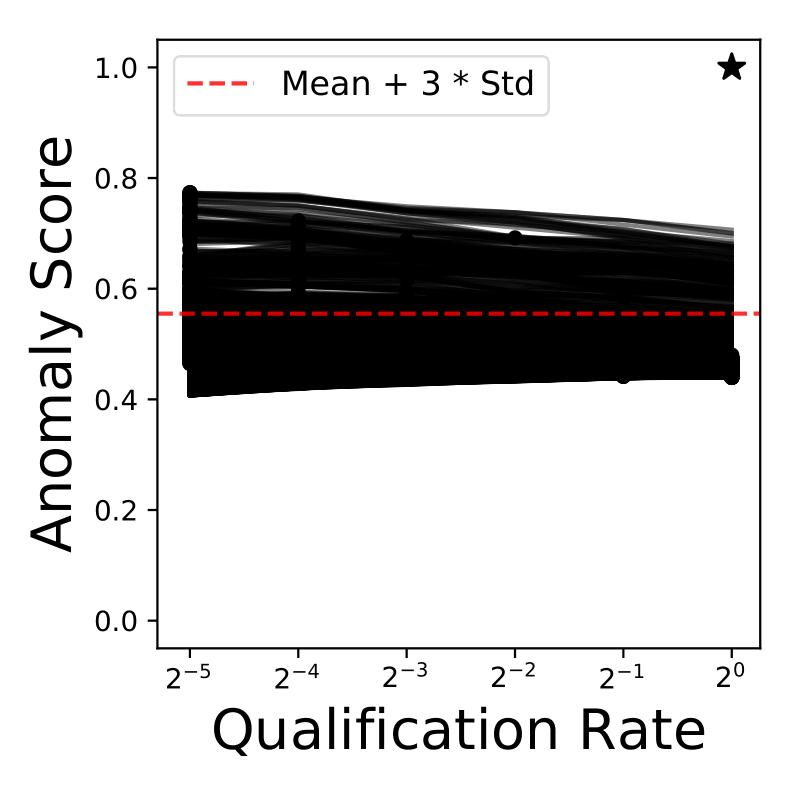}}
	\subfloat[Apex extraction]
	{\label{fig:eeg:sxray}\includegraphics[scale=0.09]{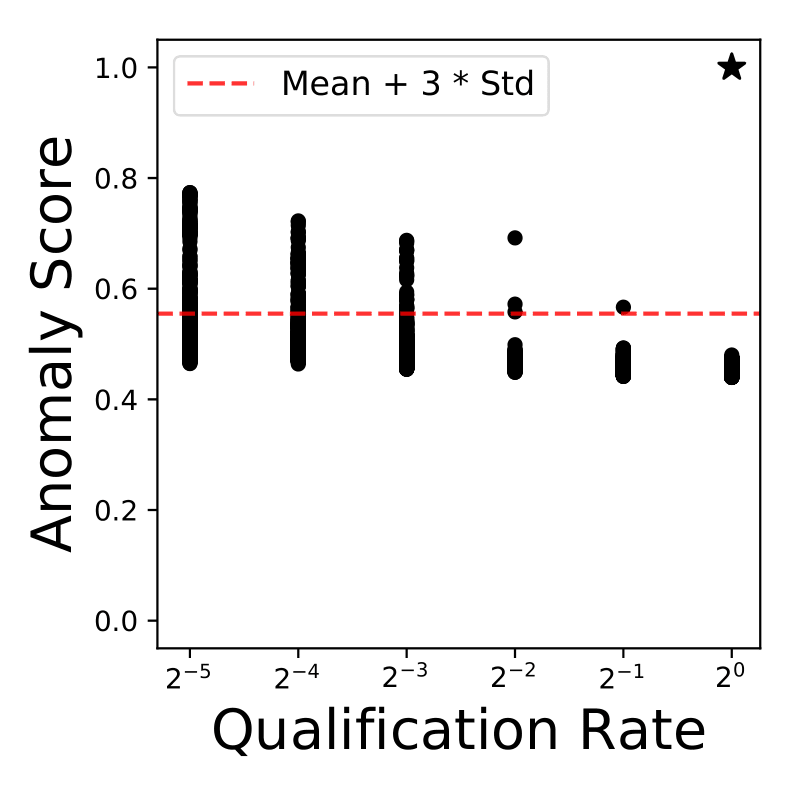}}
	\subfloat[Outlier grouping]
	{\label{fig:eeg:clust}\includegraphics[scale=0.09]{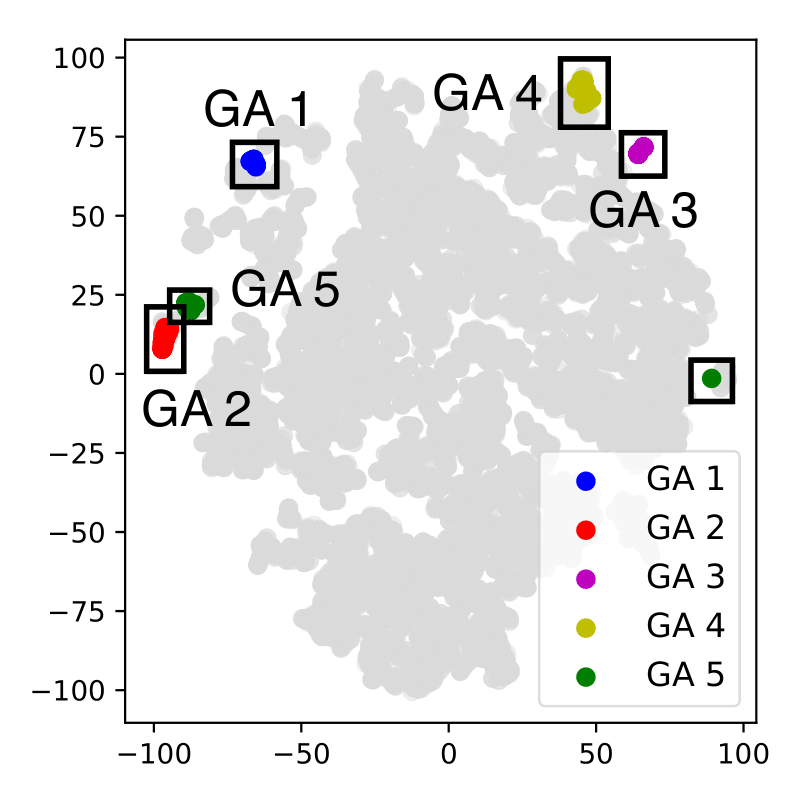}}
	\subfloat[Anomaly iso-curves]
	{\label{fig:eeg:iso}\includegraphics[scale=0.09]{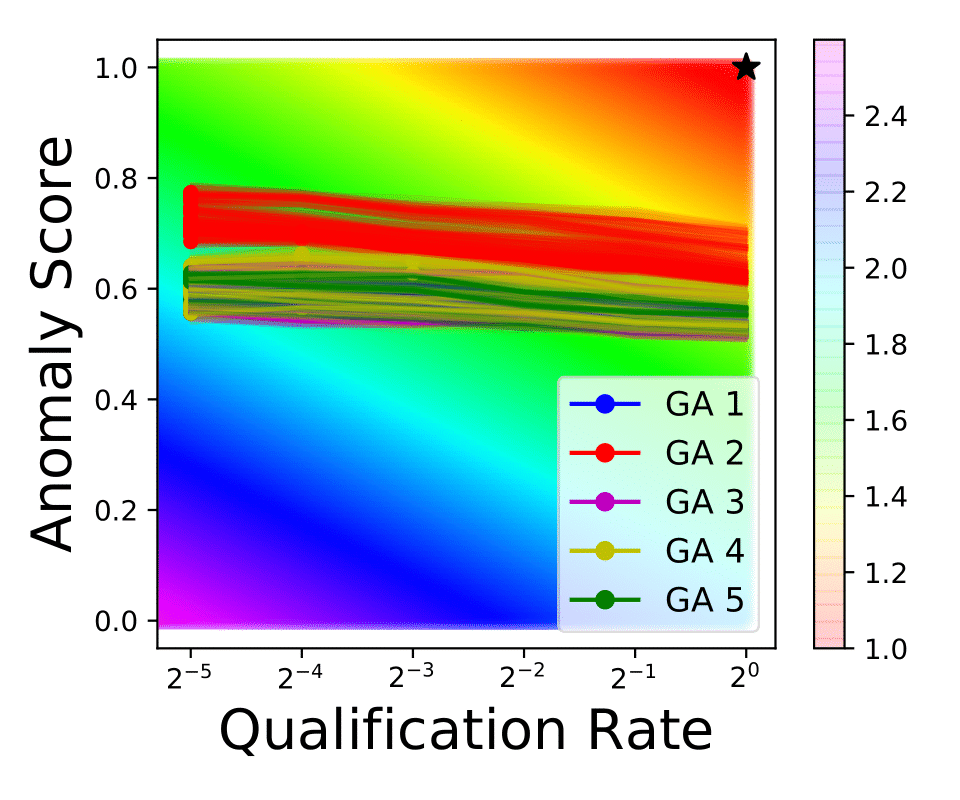}}
	\subfloat[Scoring]
	{\label{fig:eeg:score}\includegraphics[scale=0.09]{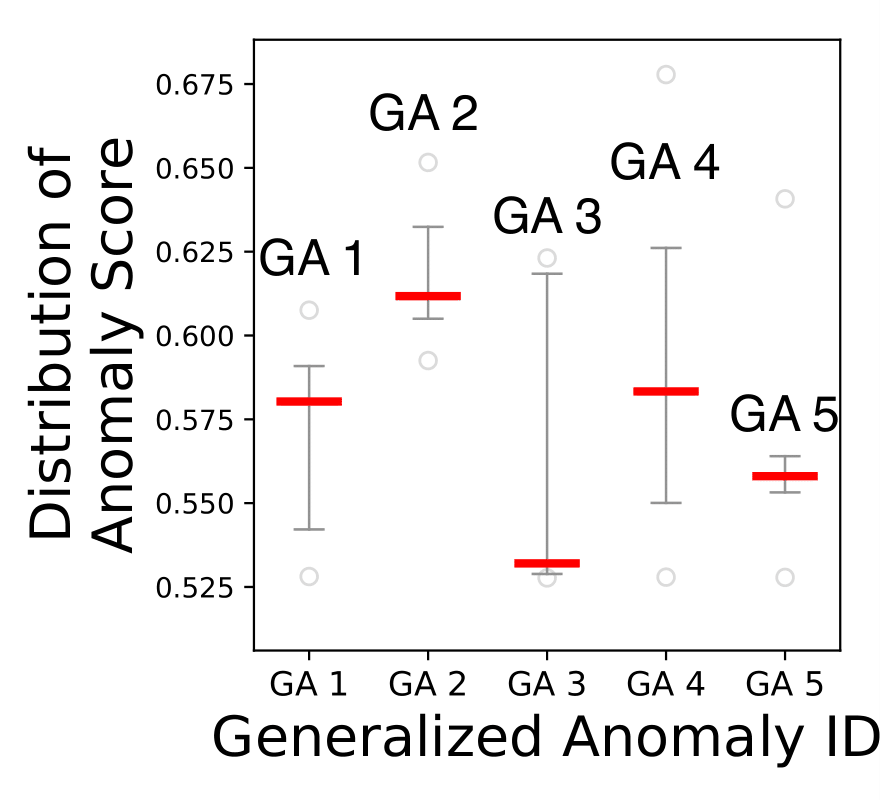}}
	\caption{\myunderline{\method works on real-world \eeg data}. Assigns highest anomaly score to \group anomaly GA2 that corresponds to seizures as we show in Figure~\ref{fig:crown2}.\label{fig:eeg:grp}}
\end{figure*}

\subsection{Point Anomalies} We compare \methodp to the following state-of-the-art ensemble baselines.

\setlist{nolistsep}
\begin{enumerate}[noitemsep]
    \item \ifr: Isolation Forest~\cite{liu2008isolation} uses an ensemble of randomized trees to flag anomalies.
    \item \loda: Lightweight on-line detector of anomalies~\cite{pevny2016loda} is a projection based histogram ensemble.
    \item \rcf: Robust Random Cut Forest~\cite{guha2016robust} are tree ensembles that use sketch based anomaly detector. 
\end{enumerate}

%
%

To evaluate the effectiveness, we compare \methodp to state-of-the-art ensemble baselines on a set of real-world point-cloud benchmark outlier detection datasets.
We use average precision (AP) and receiver operating characteristic (ROC) scores as our evaluation metrics. We plot the scores (AP and ROC score) for each competing method on all the benchmark datasets in  Figure~\ref{fig:aproc}. 

If the points are below the 45 degree line where each point represents a dataset listed in Table~\ref{tab:bmdatatable}, then it indicates that \methodp outperforms the competition in those datasets. As shown in Figure~\ref{fig:aproc}, for both the evaluation metrics, \methodp beats or at least ties with all baselines on majority of the datasets~(see Figure~\ref{fig:crown1}). The quantitative evaluation demonstrates that \methodp is superior to its competitors in terms of evaluation performance as well as obeys all the proposed axioms while none of the competition obeys the axioms.\looseness=-1

\balance

\subsection{Group Anomalies}
We evaluate the effectiveness of \method on real-world intrusion dataset that has attributes describing duration of attack, source and destination bytes. Note that we do not include \group anomaly detection methods for comparison as they require group structure information, hence do not apply to our setting. Figure~\ref{fig:http:hex} shows source bytes plotted against destination bytes for the points. 
Figures~\ref{fig:http:xray} -- \ref{fig:http:gs} shows the \xray plot with scores trajectory, \sxray with candidate points above the threshold (set at mean + 3 standard deviation of scores in full dataset), identified \group{s} and the \genanomaly score for each detected \group. \method matches ground truth as it detects the three anomalous \group{s} as shown in Figure~\ref{fig:http:post}. In short \method is able to detect \group{s} that correspond to distributed-denial-of-service attack.
\begin{figure}
	\centering
	\subfloat[]{\includegraphics[width=1.7in]{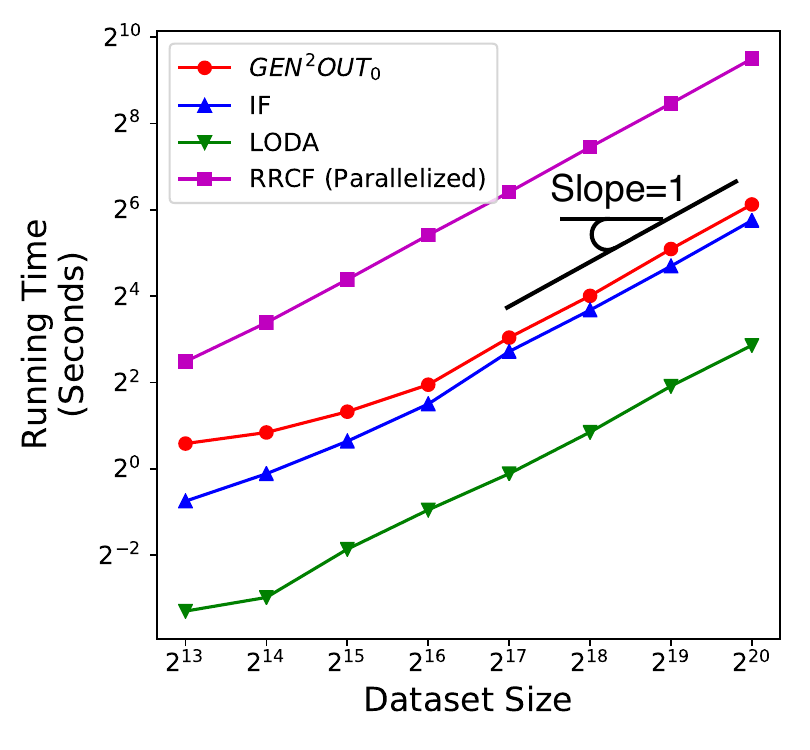} \label{fig:scalability1}} 
	\subfloat[]{\includegraphics[width=1.7in]{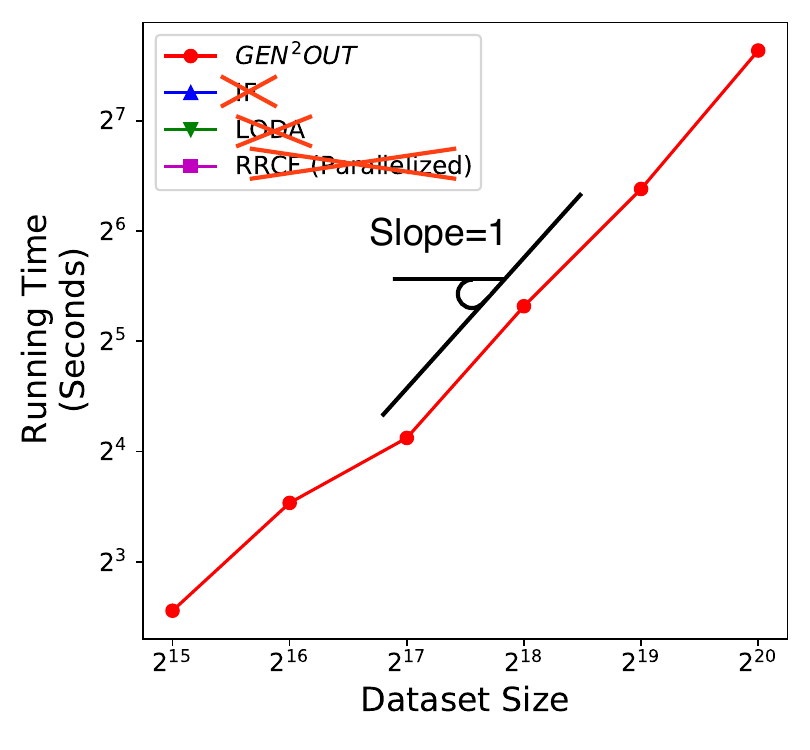} \label{fig:scalability2}}
	\caption{\label{fig:gen:scale}(a) \myunderline{\methodp is fast and scalable}:
		Evaluation on benchmark datasets show that \method (in red) scales linearly (eventual slope=1 in log-log scales). Note that none of the competitors obeys the axioms, and \rcf is significantly slower. (b) \myunderline{\method is fast and scalable}, linear in size of input.}
\end{figure}

\subsection{Scalability}
To quantify the scalability, we empirically vary the number of observations in the chosen dataset and plot against the wall-clock running time (on 3.2 GHz 36 core CPU with 256 GB RAM) for the methods. First we compare \methodp against the competitors in Figure~\ref{fig:scalability1} for \ptanomalies. The running time curve of \methodp is parallel to the running time curve of \ifr, which shows that \methodp does not increase time complexity except adding a small constant overhead for estimating the depth function $H(.)$. The running time of RRCF is much higher than others even after implementing the trees in parallel. Note that only \methodp obeys the axioms.
For \genanomalies, Figure~\ref{fig:scalability2} reports the wall-clock running time of \method as we vary the data size. Notice that \method scales linearly with input size. Importantly, competitors do not apply as they require additional information.

\section{\method at Work}

\subsection{No False Alarms.} When applied to datasets containing only normal groups that are relatively equal in size, \method correctly identifies them as normal groups i.e. does not flag any set of points as anomalous \group. To illustrate this phenomenon, we apply \method to \optdigits dataset which contains the feature representation of numerical digits.
\begin{figure}
	\centering
	\subfloat[Data heatmap]
	{\label{fig:opt:hex}\includegraphics[scale=0.15]{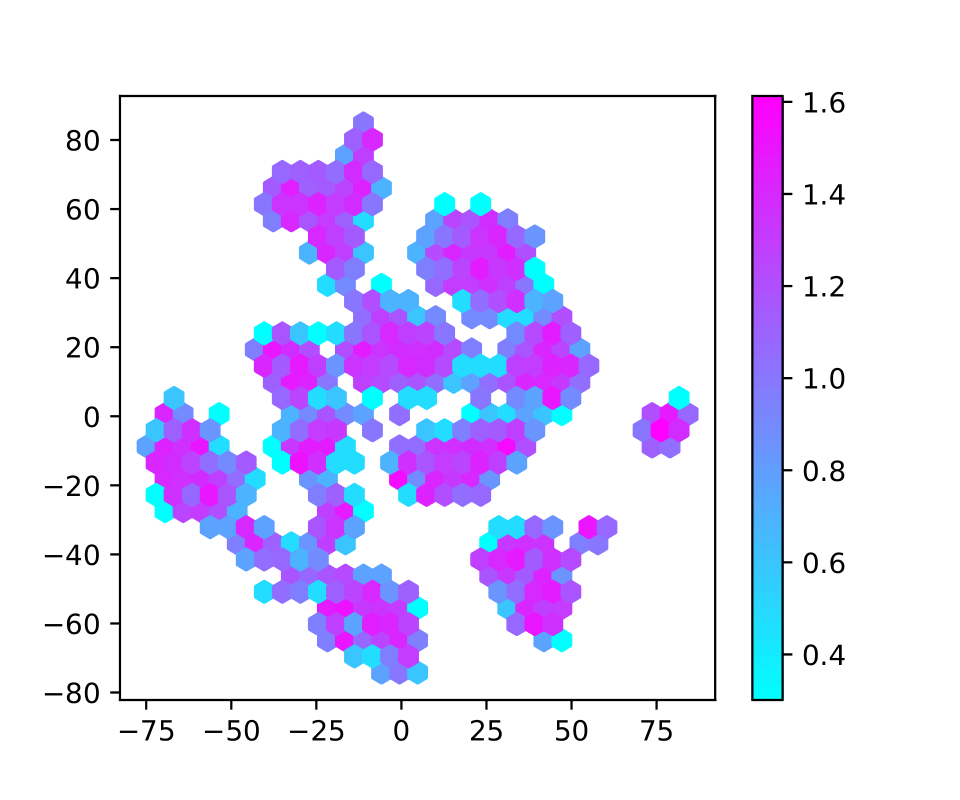}}
	\subfloat[X-Ray plot]
	{\label{fig:opt:xray}\includegraphics[scale=0.14]{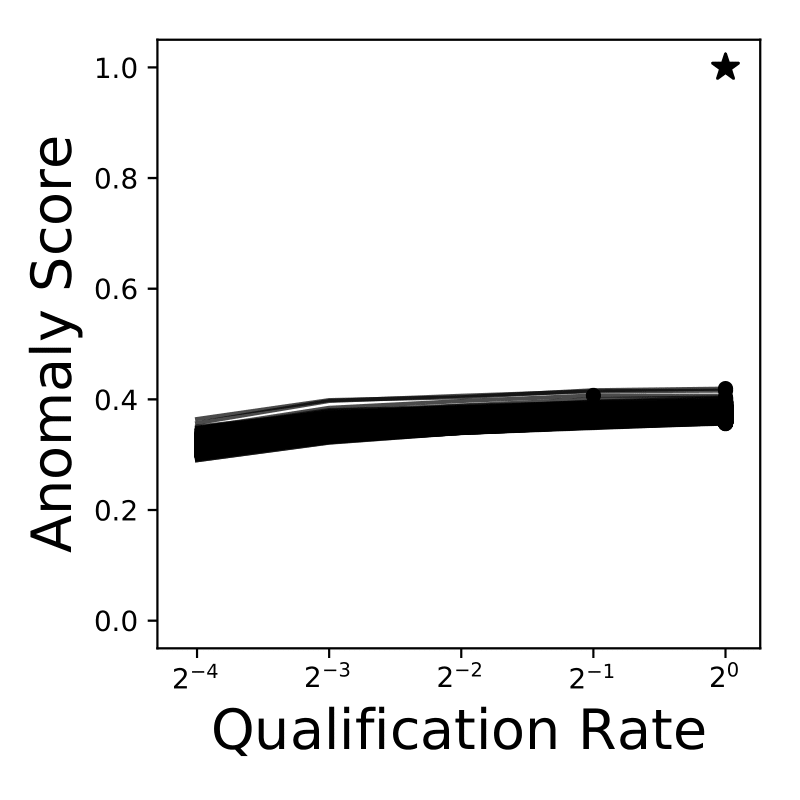}}
	\caption{\myunderline{\method works}. It correctly flags no anomalies in the \optdigits dataset}
\end{figure}

To better visualize the dataset, we embed the points in two dimensional space using tSNE~\cite{van2008visualizing} as shown in Figure~\ref{fig:opt:hex}. 
It is a balanced dataset, where we have equal number of points for each digit, hence no \group is present. \xray plot (Figure~\ref{fig:opt:xray}) shows that all the score trajectories are below $0.5$ (scores close to $1$ are anomalous) with mean score at $0.36$ in full dataset. Hence, we do not find \group and correctly so.\looseness=-1

\subsection{Attention Routing in Medicine.}
\label{subsec:discovery}

We apply \method on EEG recordings for the epileptic patient (PT1) -- PT1 suffered through onset of two seizures in our recording clips; our motivating application. We extract four simple statistical measures from the subsequences of the time series features, namely mean, variance, skewness and kurtosis, by sliding a thirty minute window with two minutes overlap. Figure~\ref{fig:eeg:heat} shows $2-$dimensional tSNE representation of the data.

We then compute the \genanomaly scores over time (within each window) for each detected \group. Since the scores generated by \method are comparable, we draw attention to the most anomalous time point, where the seizures occurred as the detected \group{s} correspond to seizure time period. The steps of \method are illustrated in Figure~\ref{fig:eeg:grp} when applied to multi-variate \eeg data. Note that we find, several \group{s} as shown in Figure~\ref{fig:eeg:grp}. Of the detected \group{s}, the \group receiving highest score (GA2) is plotted over the raw voltage recordings over time for the patient. The \group corresponds to the ground truth seizure duration (see Figure~\ref{fig:crown2}). These time points that we direct attention to would assist the domain expert (in this case a clinician) in decision making by alleviating cognitive load of examining all time points.

\section{Conclusions}
\label{sec:concl}
We presented \method{} -- a principled anomaly detection algorithm 
that has the following properties.

\bit
\item {\bf \Universal:} We propose five axioms
that
\method obeys them, in contrast to top competitors.

\item {\bf \algo:} Propose \dgen{} -- simultaneously detects point and \group anomalies -- \method.  It does not require information on group structure, and ranks detected \group{s} of varying sizes in order of their anomalousness.

\item 
{\bf \scale:}
Linear on the input size; requires minutes on 1M dataset
on a stock machine.
\item{\bf \effective}:
Applied on real-world data~(see Figure~\ref{fig:crown} and \ref{fig:aproc}),
\method wins in most cases over $27$ benchmark datasets for point anomaly detection, 
and agrees with ground truth on seizure detection as well as \group detection tasks.
\eit

\noindent {\bf Reproducibility}: Source code for algorithms are publicly available  at \url{https://github.com/mengchillee/gen2Out}.

\looseness=-1

\section*{Acknowledgments}
This material is based upon work
supported by the National Science Foundation
under Grant No. 1632891
(BRAIN initiative NSF EPSCoR RII-2 FEC OIA). This work is supported in part by NSF CAREER 1452425, and also by the Pennsylvania Infrastructure Technology Alliance, a partnership of Carnegie Mellon, Lehigh University and the Commonwealth of Pennsylvania’s Department of Community and Economic Development (DCED).
The project AIDA - Adaptive, Intelligent and Distributed Assurance Platform (reference POCI-01-0247-FEDER-045907) leading to this work is co-financed by the ERDF - European Regional Development Fund through the Operacional Program for Competitiveness and Internationalisation - COMPETE 2020 and by the Portuguese Foundation for Science and Technology - FCT under CMU Portugal.
Any opinions, findings, and conclusions or recommendations expressed in this
material are those of the author(s) and do not necessarily reflect the views
of NSF, the U.S. Government or other funding parties.
The U.S. Government is authorized to reproduce and
distribute reprints for Government purposes notwithstanding
any copyright notation here on.

\clearpage
\bibliographystyle{abbrv}
\bibliography{BIB/paper.bib}

\begin{thebibliography}{10}

\bibitem{aggarwal2015outlier}
C.~C. Aggarwal.
\newblock Outlier analysis.
\newblock In {\em Data mining}, pages 237--263. Springer, 2015.

\bibitem{10.5555/44935.44950}
M.~F. Barnsley and A.~D. Sloan.
\newblock A better way to compress images.
\newblock {\em BYTE}, 13(1):215–223, Jan. 1988.

\bibitem{blazquez2021review}
A.~Bl{\'a}zquez-Garc{\'\i}a, A.~Conde, U.~Mori, and J.~A. Lozano.
\newblock A review on outlier/anomaly detection in time series data.
\newblock {\em ACM CSUR}, 54:1--33, 2021.

\bibitem{boukerche2020outlier}
A.~Boukerche, L.~Zheng, and O.~Alfandi.
\newblock Outlier detection: Methods, models, and classification.
\newblock {\em ACM Computing Surveys (CSUR)}, 53(3):1--37, 2020.

\bibitem{breunig2000lof}
M.~M. Breunig, H.-P. Kriegel, R.~T. Ng, and J.~Sander.
\newblock Lof: identifying density-based local outliers.
\newblock In {\em ACM SIGMOD}, pages 93--104, 2000.

\bibitem{chakravarthy2009controlling}
N.~Chakravarthy, S.~Sabesan, K.~Tsakalis, and L.~Iasemidis.
\newblock Controlling epileptic seizures in a neural mass model.
\newblock {\em Journal of Combinatorial Optimization}, 17(1):98--116, 2009.

\bibitem{chalapathy2018group}
R.~Chalapathy, E.~Toth, and S.~Chawla.
\newblock Group anomaly detection using deep generative models.
\newblock In {\em ECML-PKDD}, pages 173--189, 2018.

\bibitem{chandola2009anomaly}
V.~Chandola, A.~Banerjee, and V.~Kumar.
\newblock Anomaly detection: A survey.
\newblock {\em ACM computing surveys (CSUR)}, 41(3):1--58, 2009.

\bibitem{emmott2013systematic}
A.~F. Emmott, S.~Das, T.~Dietterich, A.~Fern, and W.-K. Wong.
\newblock Systematic construction of anomaly detection benchmarks from real
  data.
\newblock In {\em KDD - ODD workshop}, pages 16--21, 2013.

\bibitem{guha2016robust}
S.~Guha, N.~Mishra, G.~Roy, and O.~Schrijvers.
\newblock Robust random cut forest based anomaly detection on streams.
\newblock In {\em ICML}, pages 2712--2721, 2016.

\bibitem{gupta2013outlier}
M.~Gupta, J.~Gao, C.~C. Aggarwal, and J.~Han.
\newblock Outlier detection for temporal data: A survey.
\newblock {\em IEEE TKDE}, 26(9):2250--2267, 2013.

\bibitem{he2003discovering}
Z.~He, X.~Xu, and S.~Deng.
\newblock Discovering cluster-based local outliers.
\newblock {\em Pattern Recognition Letters}, 24(9-10):1641--1650, 2003.

\bibitem{hutson2018predictability}
T.~Hutson, D.~Pizarro, S.~Pati, and L.~D. Iasemidis.
\newblock Predictability and resetting in a case of convulsive status
  epilepticus.
\newblock {\em Frontiers in neurology}, 9:172, 2018.

\bibitem{kriegel2008angle}
H.-P. Kriegel, M.~Schubert, and A.~Zimek.
\newblock Angle-based outlier detection in high-dimensional data.
\newblock In {\em KDD}, pages 444--452, 2008.

\bibitem{krishnan2015epileptic}
B.~Krishnan, I.~Vlachos, Z.~Wang, J.~Mosher, I.~Najm, R.~Burgess, L.~Iasemidis,
  and A.~Alexopoulos.
\newblock Epileptic focus localization based on resting state interictal meg
  recordings is feasible irrespective of the presence or absence of spikes.
\newblock {\em Clinical Neurophysiology}, 126(4):667--674, 2015.

\bibitem{krizhevsky2012imagenet}
A.~Krizhevsky, I.~Sutskever, and G.~E. Hinton.
\newblock Imagenet classification with deep convolutional neural networks.
\newblock {\em Advances in neural information processing systems},
  25:1097--1105, 2012.

\bibitem{liu2008isolation}
F.~T. Liu, K.~M. Ting, and Z.-H. Zhou.
\newblock Isolation forest.
\newblock In {\em ICDM}, pages 413--422. IEEE, 2008.

\bibitem{liu2012isolation}
F.~T. Liu, K.~M. Ting, and Z.-H. Zhou.
\newblock Isolation-based anomaly detection.
\newblock {\em ACM Transactions on Knowledge Discovery from Data (TKDD)},
  6(1):1--39, 2012.

\bibitem{muandet2013one}
K.~Muandet and B.~Sch{\"o}lkopf.
\newblock One-class support measure machines for group anomaly detection.
\newblock {\em arXiv preprint arXiv:1303.0309}, 2013.

\bibitem{pevny2016loda}
T.~Pevn{\`y}.
\newblock Loda: Lightweight on-line detector of anomalies.
\newblock {\em Machine Learning}, 102(2):275--304, 2016.

\bibitem{Rayana:2016}
S.~Rayana.
\newblock Odds library, 2016.

\bibitem{schubert2017dbscan}
E.~Schubert, J.~Sander, M.~Ester, H.~P. Kriegel, and X.~Xu.
\newblock Dbscan revisited, revisited: why and how you should (still) use
  dbscan.
\newblock {\em ACM TODS}, 42:1--21, 2017.

\bibitem{shorvon2009epilepsy}
S.~Shorvon.
\newblock {\em Epilepsy}.
\newblock Oxford Neurology Library. OUP Oxford, 2009.

\bibitem{shyu2003novel}
M.-L. Shyu.
\newblock A novel anomaly detection scheme based on principal component
  classifier.
\newblock In {\em Proc. ICDM Foundation and New Direction of Data Mining
  workshop, 2003}, pages 172--179, 2003.

\bibitem{toth2018group}
E.~Toth and S.~Chawla.
\newblock Group deviation detection methods: a survey.
\newblock {\em ACM CSUR}, 51:1--38, 2018.

\bibitem{tsakalis2006control}
K.~Tsakalis and L.~Iasemidis.
\newblock Control aspects of a theoretical model for epileptic seizures.
\newblock {\em International Journal of Bifurcation and Chaos},
  16(07):2013--2027, 2006.

\bibitem{van2008visualizing}
L.~Van~der Maaten and G.~Hinton.
\newblock Visualizing data using t-sne.
\newblock {\em JMLR}, 9, 2008.

\bibitem{vlachos2016concept}
I.~Vlachos, B.~Krishnan, D.~M. Treiman, K.~Tsakalis, D.~Kugiumtzis, and L.~D.
  Iasemidis.
\newblock The concept of effective inflow: application to interictal
  localization of the epileptogenic focus from ieeg.
\newblock {\em IEEE Transactions on Biomedical Engineering}, 64(9):2241--2252,
  2016.

\bibitem{xiong2011hierarchical}
L.~Xiong, B.~P{\'o}czos, J.~Schneider, A.~Connolly, and J.~VanderPlas.
\newblock Hierarchical probabilistic models for group anomaly detection.
\newblock In {\em AISTATS}, pages 789--797, 2011.

\bibitem{yu2015glad}
R.~Yu, X.~He, and Y.~Liu.
\newblock Glad: group anomaly detection in social media analysis.
\newblock {\em ACM Transactions on Knowledge Discovery from Data (TKDD)},
  10(2):1--22, 2015.

\end{thebibliography}

\end{document}